\documentclass{article}

\usepackage[final]{neurips_2024}
\usepackage[nohints]{minitoc}

\usepackage{wrapfig}

\usepackage{bm,bbm}
\usepackage{pifont}
\usepackage{adjustbox}
\usepackage{verbatim}
\usepackage{amsfonts}
\usepackage{amsmath,amssymb}
\usepackage{amsthm}
\usepackage{algorithm}
\usepackage{algorithmic}
\usepackage{mathtools}
\usepackage{thmtools} 
\usepackage{thm-restate}
\usepackage{enumitem}
\usepackage{graphicx}
\usepackage{float}
\usepackage{setspace}
\usepackage{natbib}

\usepackage{cancel}
\usepackage{tikz}
\usepackage{subcaption}
\usepackage{placeins}
\usepackage{float}
\usetikzlibrary{patterns}
\usepackage{tabularx}

\usepackage[utf8]{inputenc}

\usepackage{verbatim}
\usepackage{todonotes}
\usepackage{algorithm}

\usepackage{nicefrac} 
\usepackage{booktabs}   
\usepackage{microtype}
\usepackage{url,hyperref}

\declaretheorem[name=Theorem,numberwithin=section]{theorem}
\declaretheorem[name=Lemma,numberwithin=section]{lemma}

\declaretheorem[name=Definition,numberwithin=section]{definition}
\declaretheorem[name=Example,numberwithin=section]{exmp}
\declaretheorem[name=Assumption,numberwithin=section]{assmpt}

\newcommand{\eqdef}{\ensuremath{\triangleq}}

\usepackage{longtable,tcolorbox}

\newcommand{\as}[1]{{\color{violet}AS: #1}}

\newcommand{\cZ}{\mathcal{Z}}

\def\cC{\mathcal{C}}
\def\cD{\mathcal{D}}
\def\cE{\mathcal{E}}
\def\cF{\mathcal{F}}
\def\cG{\mathcal{G}}
\def\cH{\mathcal{H}}
\def\cI{\mathcal{I}}

\def\cL{\mathcal{L}}
\def\cM{\mathcal{M}}
\def\cN{\mathcal{N}}

\def\cP{\mathcal{P}}

\def\cT{\mathcal{T}}
\def\cU{\mathcal{U}}
\def\cV{\mathcal{V}}

\def\ones{\boldsymbol{1}}

\def\mB{\mathbb{B}}

\def\mE{\mathbb{E}}

\def\mN{\mathbb{N}}

\def\mP{\mathbb{P}}

\def\mR{\mathbb{R}}

\def\cP{\mathcal{P}}

\def\notdom{\not\preceq_{\cC}} 
\def\vectle{\preceq_{\cC}} 
\def\vectl{\prec_{\cC}} 
\def\vectdom{\prec_{\cC\setminus\{0\}}}
\def\vectneq{\npreceq_{\cC}}
\def\pftrue{\mathcal{P}^{\ast}}
\def\pfestm{\hat{\cP}_t}
\def\klobjl{\kl{z^{\top}\mklestm}{z^{\top}\mkl}}
\def\cD{\mathcal{D}}
\def\numarmkt{N_{k,t}}
\def\armset{[K]}
\def\pfestmstop{\hat{\mathcal{P}}_{\tau}}
\def\paretospace{\mathcal{Z}}

\def\goodevent{\mathcal{G}_{T}}
\def\eqdef{\triangleq}
\def\mklestm{\hat{M}_{k,t}}
\def\mkl{M_{k}}

\def\badevent{\bar{\cG}_{T}}
\def\concevent{\cE}
\def\cone{\mathcal{C}}
\def\stoptime{\tau}
\def\chartime{\cT_{M,\cC}}
\def\confball{c(t,\delta)}
\def\mestm{\hat{M}_{t}}

\def\instset{\mathcal{M}}

\def\pspace{\mathcal{Z}}

\usepackage{todonotes}

\newcommand{\ind}[1]{\mathbbm{1}_{#1}}
\newcommand{\hdist}[2]{d_{\mathrm{H}}(#1,#2)}
\newcommand{\pdist}[2]{d_{\mathrm{P}}\left(#1,#2\right)}

\newcommand{\muestm}[2]{\hat{M}_{#1,#2}}
\newcommand{\mutrue}[1]{M_{#1}}

\newcommand{\cc}[1]{\text{ch}\left({#1}\right)}
\newcommand{\kl}[2]{d_{\textrm{KL}}\left(#1,#2\right)}
\newcommand{\altset}[1]{\Lambda_{\pi^*}\left(#1\right)}

\title{Preference-based Pure Exploration}

\author{%
  Apurv Shukla\thanks{This work was done when the author was at Texas A\& M University. The author is currently at the  University of Michigan, Ann Arbor.}\\
  Department of ECE, Texas A\&M University \\
  College Station, TX 77840 \\
  \texttt{apurv.shukla@umich.edu}
  \And 
  Debabrota Basu \\
  \'{E}quipe School, Univ. Lille, Inria, CNRS\\
  Centrale Lille, UMR-9189 - CRIStAL, France\\  \texttt{debabrota.basu@inria.fr}
}

\begin{document}
\maketitle
\doparttoc
\faketableofcontents 

\begin{abstract}
We study the preference-based pure exploration problem for bandits with vector-valued rewards. The rewards are ordered using a (given) preference cone $\cC$ and our goal is to identify the set of Pareto optimal arms. First, to quantify the impact of preferences, we derive a novel lower bound on sample complexity for identifying the most preferred policy with a confidence level $1-\delta$. Our lower bound elicits the role played by the geometry of the preference cone and punctuates the difference in hardness compared to existing best-arm identification variants of the problem. We further explicate this geometry when the rewards follow Gaussian distributions. We then provide a convex relaxation of the lower bound and leverage it to design the Preference-based Track and Stop (PreTS) algorithm that identifies the most preferred policy. Finally, we show that the sample complexity of PreTS is asymptotically tight by deriving a new concentration inequality for vector-valued rewards.
\end{abstract}

\section{Introduction}
Following COVID-19, the importance of reliable clinical trials and corresponding data acquisition to design effective drugs has gained wider recognition.
However, conducting large-scale clinical trials is cost and time intensive as it requires working with large number of patients and following up their medical conditions over time.
In the past two decades, this has led to doubling in the cost to bring a drug to the market, i.e., to $\$ 2.6$ billion with a 12-year drug development horizon and 90\% failure rate during the clinical trial~\citep{mullard2014new,SUN20223049}.
However, due to the rise of systematic data acquisition about biological systems,  pharmaceutical firms are interested in harvesting the collected data for drug discovery~\citep{gaulton2012chembl,reker2015active}. 
Thus, machine learning-based methods are increasingly studied and deployed as a promising avenue for identifying potentially successful drugs with less patient involvement, increasing the ``hit rate", and speeding up the development process~\citep{jayatunga2022ai,smer2023discovery,sadybekov2023computational,hasselgren2024artificial}.
But deciding whether a drug is successful depends on multiple and often conflicting objectives regarding safety, efficacy, and pharmacokinetic constraints~\citep{lizotte2016multi}.
For example, COV-BOOST~\citep{munro2021safety} demonstrates a phase II vaccine clinical trial conducted on 2883 participants to measure the immunogenicity indicators (e.g. cellular response, anti-spike IgG and NT$_{50}$) of different Covid-19 vaccines as a booster (third dose).
Experts decide how different indicators are preferred over one another, and above different thresholds~\citep{jayatunga2022ai}.
This motivates us to study a sequential decision-making problem, where we aim to conduct minimum number of experiments to acquire informative data, 
and to reliably validate a hypothesis with multiple objectives by imposing preferences over them.

Problems of such nature can be modeled as a multi-armed bandit (in brief, bandits), which is an established framework for sequential decision-making under uncertainty~\citep{lattimore_szepesvari_2020}.
In bandits, a learner has access to an instance of $K$ decisions (or arms).
Each arm $k \in \{1,\ldots, K\}$ corresponds to a probability distribution $P_k$ of feedback (rewards) with unknown mean $\mu_k$.
At each step $t \in \mathbb{N}$, the learner interacts with the instance by taking a decision $k_t$ (analogously pulling an arm), and observes a noisy reward $R_t$ from the corresponding distribution of rewards $P_{k_t}$.
The goal of the learner is to identify the arm with the highest expected reward over a certain confidence level through minimum number of interactions with the instance. 
This is popularly known as a fixed-confidence Best Arm Identification (BAI) in bandit literature~\citep{jamieson2014best,garivier2016optimal,soare2014best}, which is a special case of pure exploration problems~\citep{even2006action,bubeck2009pure,auer2016pareto}.

The bandit literature spanning over a century mostly focuses on a scalar reward, i.e., a single objective. In our problem, each reward $R_t$ is a real-valued vector of $L \in \mathbb{N}$ objectives, and thus, the unknown mean vector of each arm $M_k \in \mathbb{R}^L$. Since the objectives can be often conflicting, there might not exist a single best arm. Rather, there exists a Pareto Optimal Set of arms~\citep{drugan2013designing,auer2016pareto}.
Given a set of preferences over the objectives, the Pareto Optimal Set consists of arms whose mean vectors dominate the mean vectors of any other arm outside the set. Keeping generality, we assume that preferences are defined by a cone of vectors $\cC \subseteq \mR^L$. Every $\cC$ induces a set of partial or incomplete orders over the $L$ objectives~\citep{jahn2009vector,lohne2011vector}. Given the preference cone $\cC$, we aim to \textit{exactly} identify the complete Pareto Optimal Set with a confidence level $(1-\delta) \in [0,1)$ using as few interactions as possible.
We refer to this problem Preference-based Pure Exploration (\textbf{PrePEx}). Recently, \cite{auer2016pareto,kone2023adaptive,kone2023bandit,garivier2024sequential} consider a special case of PrePEx, where the preference is known. To the best of our knowledge, \cite{ararat2023vector} and \cite{korkmaz2023bayesian} are the only studies of PrePEx from frequentist and Bayesian angles, respectively. Here, we consider a frequentist approach as in \citep{ararat2023vector}. 
However, their goal is to identify points that are in the Pareto Optimal Set or very close to it.
In contrast, we focus on exactly identifying the Pareto Optimal Set. 
Additionally, \cite{ararat2023vector} propose a gap-based elimination algorithm to solve the problem that generalises the algorithm of~\cite{even2006action}.
But in BAI, there is another paradigm of designing efficient algorithms that solves and tracks the exact lower bound on the expected time to identify the best arm $(1-\delta)$ correctly~\citep{garivier2016optimal,Degenne2019PureEW}.
We explore this paradigm for PrePEx and ask two questions:

\textit{What is the exact lower bound of \emph{PrePEx} for identifying the Pareto Optimal Set, and how to design a computationally tractable algorithm matching this bound?}

We address them affirmatively in our contributions:\\
    \textbf{1. Lower Bound for PrePEx.} In Theorem~\ref{thm:lower-bound}, we study hardness of PrePEx problems by deriving the novel lower bound on the expected sample complexity of any algorithm to yield the exact Pareto Optimal Set with confidence $(1-\delta)$. The challenge here is to extend the classical BAI lower bound~\citep{garivier2016optimal} to a set of confusing instances given $\cone$. We observe that unlike BAI, distinguishability of two arms in PrePEx depends on their projections on the cone polar to $\cC$. We also show that our lower bound generalises the lower bound for pure exploration under known constraints~\citep{carlsson2024pure}. Additionally, we provide an exact characterization the lower bound further for Gaussian reward distributions in Theorem~\ref{thm:gaussian-bandit}. It shows that the hardness depends on the bilinear projection of the mean matrix of arms onto the boundary of a normal cone of policies and the preferences. This is novel w.r.t. the existing gap-dependent lower bounds that hold either for a narrow range of $\mu_a$'s~\citep{ararat2023vector}, or fixed preference~\citep{kone2023adaptive}.\\
    \textbf{2. Algorithm Design.} First, we observe that the optimisation problem in our lower bound involves minimisation over a non-convex set. We provide a convex relaxation of the problem based on ideas from disjunctive programming (Theorem~\ref{thm:ccl-alt-set} and~\ref{thm:perturbation-properties}). We then leverage this lower bound to propose a novel Track-and-Stop~\citep{garivier2016optimal} style algorithm, called PreTS (Preference-based Track-and-Stop). In Theorem~\ref{thm:stopping-time}, we devise a new stopping rule that can handle the preference-aligned suboptimality gaps between the arms.\\
    \textbf{3. Sample Complexity Analysis.} Finally, we provide an upper bound on sample complexity of PreTS. This requires us to define a distance metric between two pareto sets of arms, and proving a concentration bound with respect to this metric (Theorem~\ref{thm:mean-concentration}). In Theorem~\ref{thm:upper-tns}, we prove that sample complexity of PreTS matches the convexified lower bound up to constants.
 
\subsection{Related Works}\vspace*{-.5em}
In the past decade, works on multi-armed bandits also focuses on  pure-exploration in addition to regret minimization. Regret minimization and pure-exploration differ in the sense when arms in pure-exploration are immediately discarded upon being deemed as sub-optimal, whereas, in the regret minimization setting, sub-optimal arms may still be played since they provide additional information about other arms. Pure-exploration problems have been considered in two settings: fixed-budget and fixed-confidence. The fixed-budget setting aims at bounding the probability of underestimating the best arm given a budget of samples. \cite{audibert2010best} propose the first algorithm for the fixed budget setting. Here, the budget is divided into $K-1$ rounds and at the end of every round, the arms with the lowest empirical mean are discarded. On the other hand, best-arm identification is a version of the pure-exploration problem with scalar rewards~\citep{even2006action}. In this setting, we are given a $\delta \in (0,1)$ and the goal is to identify the best-arm with probability at least $1-\delta$.
Several strategies such as those based on elimination, adaptivity, racing, upper-confidence bounds have been proposed to minimize the number of expected pulls of an arm in the fixed confidence setting by~\citep{kalyanakrishnan2012pac,gabillon2012best,jamieson2014lil,garivier2016optimal,jedra2020optimal}. Arm rewards can be modeled as a vector with
Gaussian Process~\citep{zuluaga2016pal}, linear rewards~\citep{drugan2013designing,lu2019multi}, and non-parametric rewards~\citep{turgay2018multi}, which can include contextual bandit formulations~\citep{tekin2017multi,shukla2022optimization}. In recent past, the pure exploration techniques have been successfully applied in hyperparameter tuning~\citep{li2018hyperband} and black-box optimization problems~\citep{contal2013parallel,wang2021demonstrating,wang2022procrastinated} demonstrating considerable performance gains.

In a marked deviation, given an instance of the bandit problem, the goal of this paper is to identify the entire Pareto front. A key observation in this regard is that there might be arms, which are sub-optimal for almost every objective but still lie on the Pareto front. Further, since sampling an arm returns a vector of rewards determining an arm-strategy that reduces the uncertainty in the estimate of every reward function is challenging. An immediate consequence of these differences is the fact that the complexity of identifying the Pareto front is different from that of best arm identification. ~\cite{auer2016pareto} consider the Pareto front identification problem in the multi-armed bandit model and establish sample complexity bounds for the problem in terms of relevant problem parameters in the fixed-confidence setting. The multi-armed bandit problem is further studied under cone-based preferences by~\cite{ararat2023vector}. The main contribution of~\citep{ararat2023vector} are bounds on the sample complexity of the problem in terms of gap-based notions that depend on the cone. \cite{karagozlu2024learning} builds upon this work to introduce adaptive elimination based algorithms for learning the Pareto front under incomplete preferences. When the reward vectors are Gaussian processes \cite{korkmaz2023bayesian} propose an elimination based algorithm based for identifying the Pareto front. The goal in these works is to identify the set of arms that are $\epsilon$ close to the Pareto front 
as the sample complexity to identify the exact Pareto set can be very large. \cite{kone2023adaptive} consider the problem of identifying a relevant subset of the Pareto set using a single sampling strategy Adaptive Pareto Exploration, along with different stopping rules to consider variations of the Pareto Set Identification problem. ~\cite{garivier2024sequential} consider the exact Pareto front identification problem in the multi-armed bandit setting but with fixed and known preference cone. They propose a lower bound and a computationally efficient gradient-based algorithm to implement a track-and-stop based strategy. To the best of our knowledge, ours is the first work to consider \textit{the exact Pareto front identification problem from a pure-exploration perspective}. {Therefore, our proposed framework can be used for identifying the Pareto front given a preference cone for several variants of the bandit problem including the standard multi-armed bandit problem, linear bandits, etc.}

\vspace*{-1em}\section{Preference-based Pure Exploration Problem}\vspace*{-.5em}
In this section, we formalise the fixed-confidence setting of preference-based pure exploration and introduce the notations.

\textbf{Notations.} For $n \in \mathbb{N}$, let $[n]$ denote the set $\{1,2,\ldots,n\}$. We use $\Vert \cdot \Vert_{1}, \Vert \cdot \Vert_{2}, \Vert \cdot \Vert_{\infty}$ to denote the $\ell_{1}$-norm, $\ell_2$-norm and $\ell_{\infty}$-norm, respectively. For a vector $z,\ z^{(\ell)}$ denotes its $\ell^{th}$ component. Let $e_{\ell}$ denote the vector with $1$ in the $\ell^{th}$ position and zero otherwise. $\Delta_K$ denotes the simplex on $[K]$. $\kl{P}{Q}$ measures the KL-divergence between distributions $P$ and $Q$. $\mathrm{vect}(A)$ is the vectorized version of matrix $A$. $\ones$ is the vector of all $1$'s. Further details of notations are deferred to Appendix~\ref{app:notations}.

\textbf{PrePEx: Problem Formulation.} In PrePEx, a learner can access a bandit instance with $K$ arms. Each arm $k \in [K]$ corresponds to a reward distribution $P_k$ over $\mR^L$ with unknown mean $M_k \in \mR^{L}$ and known covariance $\Sigma = \mathrm{Diag}(\sigma_1^2,\ldots,\sigma_L^2)$. Here, $L$ denotes the number of objectives corresponding to each arm.  Thus, a bandit instance can be specified with the vector of mean rewards $\{M_k\}_{k=1}^{K}$. 
For brevity, we represent them with a matrix $M \in \mR^{L \times K}$ such that its $k^{\mathrm{th}}$ column is $M_k$. 
At each time $t \in \mathbb{N}$, the learner pulls an arm $k_t \in [K]$ and observes the corresponding reward vector $R_t$ sampled from $\nu_{k_t}$. In pure exploration, the learner typically focuses on finding the best arm, i.e. the arm with highest mean~\citep{garivier2016optimal}. 
{In pure exploration, a more general setting of BAI, the learner aims to find a policy $\pi \in \Delta_K$ that dictates the arm-proportion to choose to maximize the expected reward from the instance.} 

Following the vector optimization literature~\citep{jahn2009vector,ararat2023vector}, we assume that the learner has additionally access to an ordering cone $\cC$. 
\begin{definition}[Ordering Cone]
A set $\cC \subseteq \mR^{L}$ is called a \textit{cone} if $v \in \cC$ implies that $\alpha v \in \cC$ for all $\alpha \ge 0$. A \textit{solid cone} has a non-empty interior, i.e., $\mathrm{int}(\cC) \neq \emptyset$. A \textit{pointed cone} contains the origin. A closed convex, pointed and solid cone is called an \textit{ordering cone}. 
\end{definition}
An ordering cone can be both polyhedral and non-polyhedral. Following the literature~\citep{ararat2023vector,karagozlu2024learning}, we consider access to a polyhedral ordering cone.
\begin{definition}[Polyhederal Ordering Cone]
An ordering cone $\cC$ is a polyhedral cone if $\cC \triangleq \{x \in \mR^{L} \ \vert \ Ax \ge 0\}$, where $A \in \mR^{K \times L}$ with rows $a_i^\top$. $A$ is called the half-space representation of $\cC$. 
\end{definition}
Each polyhedral ordering cone induces a set of partial order on the reward vectors in $\mR^L$. 
To ignore the redundancies and to focus on the bandit problem, we further assume that $A$ is full row-rank and $\Vert A_i\Vert_2 = 1$~\citep{ararat2023vector}. Hereafter, we call them \textit{preference cones}, and the vectors in the cone as the \textit{preferences}. We refer to \citep{jahn2009vector,lohne2011vector} for further details on cones. 
\begin{exmp}[Preference cones]
The positive orthant $\mR_+^L$ is a polyhedral ordering cone. This is the one used in the pareto-set identification literature~\citep{auer2016pareto,kone2023bandit,garivier2024sequential}. The cones with all non-negative entries are called solvency cones and used in finance~\citep{kabanov2009markets}. Another simple example is $\cC_{\pi/3} \triangleq \{(r \cos\theta, r \sin\theta) \in \mR^2 \mid r \geq 0 \wedge \theta \in [0,\pi/3]\}$, i.e.,all the 2-dimensional vectors that make an angle less than $\pi/3$ with the $x$-axis.
\end{exmp}
\begin{wrapfigure}{r}{0.35\textwidth}
\centering\vspace*{-1em}
\includegraphics[width=0.35\textwidth]{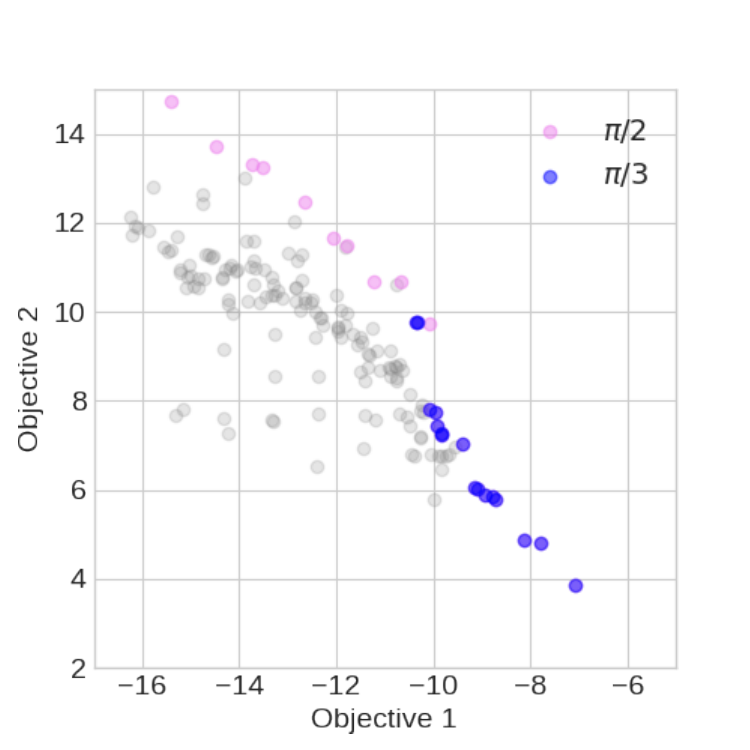}
\caption{Effect of cone selection on size of Pareto optimal set}\label{fig:eg-preference}\vspace*{-2em}
\end{wrapfigure}
\begin{definition}[Partial Order]
For every $\mu,\mu' \in \mR^{L}, \mu \vectle \mu'$ if $\mu \in \mu'+\cC$ and $\mu \vectl \mu$ if $\mu \in \mu'+\mathrm{int}(\cC)$. Alternatively, $\mu \vectle \mu'$ is equivalent to $z^{\top}(\mu - \mu') \leq 0, \forall z \in \cC$.
\end{definition}
The partial order induced by $\cC$ induces further order over the set of arms $[K]$.
\begin{definition}[Order over arms]
Consider two arms $i,j \in [K]$. (i) Arm $i$ is weakly dominated by arm $j$ iff $M_{i} \vectle M_{j}$. 
(ii) Arm $i$ dominates arm $j$ iff $M_{j}\vectdom M_{i}$. 
(iii) Arm $i$ strongly dominates arm $j$ iff $M_{j} \vectl M_{i}$.
\end{definition}
\begin{definition}[Pareto Optimal Set]
An arm $i \in \armset$ is \textit{Pareto Optimal} if it is not dominated by any other arm w.r.t. the cone $\cC$. The \textit{Pareto Optimal Set} $\cP^\ast$ is defined as the set of all Pareto Optimal arms. Let $\paretospace$ be the set of all Pareto Frontiers on $[0,1]^{K}$.
\end{definition}
Given a preference cone, a learner aims to identify exactly the Pareto Optimal Set from a finite set of arms $[K]$ whose mean rewards belong to the Pareto Optimal Set w.r.t. $\cC$. Alternatively, this vector optimization problem can be represented in the policy space as finding a policy $\pi \in \Delta_K$ supported on the Pareto Optimal Set of arms. The following vector optimization problem yields this:
\begin{equation}\label{eqn:vector-opt}
V(M) \triangleq \max_{\pi \in \Delta_K} \ M\pi \ \text{over} \ \cC.
\end{equation}
In this context, we denote the set of Pareto optimal policies as $\Pi^*(M) \triangleq \arg\max_{\pi \in \Delta_K} \ M\pi \ \text{over} \ \cC$. We assume that $\Pi^*(M)$ is non-empty.

\begin{exmp}[Pareto Optimal Sets for different cones]
    Figure~\ref{fig:eg-preference} illustrates the Pareto Optimal Sets among $2$-dimensional mean vectors of $200$ randomly selected arms under preference cones $\cC_{\pi/2}$ and $\cC_{\pi/3}$. We observe that the Pareto Optimal Sets for them (in pink and blue respectively), are completely different for the same set of arms. Thus, we have to adapt to the available preferences to solve the aforementioned problem. As noted later, the geometry of this cone plays a crucial role in determining the Pareto front.
\end{exmp}\vspace*{-.5em}

In PrePEx, we consider the problem of Equation~\eqref{eqn:vector-opt}, when the mean matrix $M$ is unknown a priori but bounded, i.e., the entries of $M,\ M_{ij} \in [M_{\min},M_{\max}]$. We denote all such mean matrices by $\cM$. Identifying the policy will lead us to identify the true Pareto front $\cP^{\ast}$. In the noisy feedback setting, the reward at time $t$ is $R_t = M_{k_t}+\eta_t$, where $\eta_t \in \mR^{L}$ is the noise vector. We assume that the noise vectors $\eta_t$ are independent of $M_{k_t}$ and also across time. Further, they are sub-Gaussian with parameter $\sigma$ and adapted to the filtration $\cF_{t}$, which is a standard assumption in the literature. 
A policy $\pi \in \Pi \subset \Delta^{K}$ is a randomized mapping from the history $\cH_{t}$ to the probability simplex over the set of arms $[K]$. \textit{In Preference-based Pure EXploration (PrePEX) problem, the goal of the learner is to identify a Pareto optimal policy in $\Pi^{\ast}$ (Equation~\eqref{eqn:vector-opt}) given an instance $M$ and a preference cone $\cC$ while observing only noisy rewards from the arms, and also using as few observations as possible.}

\begin{definition}[$(1-\delta)$-correct PrePEX] 
An algorithm for Preference-based Pure Exploration (PrePEx) is said to be $(1-\delta)$ correct if with probability $1-\delta$, it recommends a Pareto optimal policy $\pi \in \Pi^\ast$.
\end{definition}
\vspace*{-.5em}
For example, a pareto optimal policy for $\cC_{\pi/2}$ would be a distribution in $\Delta_K$ with support on the arms corresponding to the pink reward vectors (Figure~\ref{fig:eg-preference}). For $\cC_{\pi/3}$, it would be one with support on blue points. Finally, we make the standard assumption on the mean rewards. 

\begin{assmpt}[Single Parameter Exponential Family]
\label{assmpt:single-para-exp}
Let $X = (X_1,···,X_d)$ be a $d$-dimensional random vector with a distribution $P_\theta,\theta \in \Theta$. Suppose $X_1,\ldots, X_d$ are jointly continuous. The family of distributions $\{P_\theta, \theta \in \Theta\}$ is said to belong to the one parameter Exponential family if the density of $X$ may be represented in the form
$f(x\vert\theta) \eqdef h(x) \exp\left( \eta(\theta)T(x)- \psi(\theta)\right)$. We assume that the mean reward for each vector belongs to a single-parameter exponential family with variance bounded by $1$. 
\end{assmpt}

\vspace*{-.5em}\section{Lower Bound on Sample Complexity}\vspace*{-.5em}
We begin by deriving a KL-divergence based lower bound for PrePEx using techniques from~\citep{garivier2016optimal}. Our lower bound is based on establishing a change-of-measure argument in the spirit of~\citep{graves1997asymptotically,kaufmann2016complexity}. The lower bounds are derived first by defining a set of alternating instances $\Lambda$ for a given bandit instance and then by trying to compute an optimal allocation policy $w \in \Delta_K$ that maximises the sum of minimum KL-divergence between any instance in $\Lambda$ and the bandit instance under interaction. \textit{The key insight of our work is to formulate the identification of Pareto Set problem in the policy space rather than in the arm space as done in antecedent literature.} This formulation helps us to derive the KL-based lower bound, which is more general than the existing suboptimality gap-based lower bounds~\citep{auer2016pareto,ararat2023vector,garivier2024sequential}.

\textbf{The Alternating Instances with respect to Pareto Fronts.} 
The learner needs to distinguish between $M$ and any instance $\tilde{M} \in \cM \setminus \{M\}$ for which the Pareto front associated with $\tilde{M}$ is different from the one associated with $M$. At first, given an optimal policy of $M$, say $\pi^*$, it would appear that the set of confusing instances is 
$\altset{M}^{\mathrm{naive}} \eqdef \left\{\tilde{M} \in \instset: \tilde{M}\pi^\ast \vectle \max_{\pi \in \Pi} \tilde{M} \pi\right\}$.
However, this is fallacious since the instances whose rewards dominate $M$ can also confuse a policy $\pi$. Given a $\pi^\star$, the correct alternating set is the set of instances in $\cM$ whose Pareto Optimal Set is not dominated by $\pi^\star$ corresponding to $M$.
\begin{align*}
\label{eqn:alternating-set}
\altset{M} &\eqdef \left\{\tilde{M} \in \instset\setminus \{M\}: \max_{\pi \in \Pi} \tilde{M} \pi \vectneq \tilde{M}\pi^* \right\}\\
&= \left\{\tilde{M} \in \instset\setminus \{M\}: \exists z \in \cC \text{ s.t. } \max_{\pi \in \Pi} z^{\top}\tilde{M} \pi > z^{\top}\tilde{M}\pi^* \right\}\, .
\end{align*}

With this alternate set, we now establish lower bounds on the performance of any PrePEX algorithm.

\begin{theorem}[Lower Bound]
\label{thm:lower-bound}
Given a bandit model $M \in \instset$, a preference cone $\cC$, and a confidence level $\delta \in [0,1)$, the expected stopping time of any $(1-\delta)$-correct PrePEx algorithm, to identify the Pareto Optimal Set is
\begin{equation}
    \mE[\tau] \ge \cT_{M,\cC}\log\left(\frac{1}{2.4\delta}\right),
\end{equation}
where, the expectation is taken over the stochasticity of both the algorithm and the bandit instance. Here, $\cT_{M,\cC}$ is called the characteristic time of the PrePEx instance $(\cM, \cC)$ and is expressed as
\begin{equation}
\label{eqn:characteristic-time}
\left(\cT_{M, \cC}\right)^{-1} \eqdef \sup_{w \in \Delta^{K}} \inf_{\substack{\pi \in \Pi\setminus \{\pi^*\}\\\pi^* \in \Pi^*(M)}} \inf_{\tilde{M} \in \partial\altset{M}} \inf_{z \in \cC} \sum_{k=1}^{K} w_k  \kl{z^{\top}M_{k}}{z^{\top}\tilde{M}_{k}}\, ,
\end{equation}
where $\partial\altset{M} \eqdef \cup_{\Pi \setminus \{\pi^{\ast}\}} \left\{\tilde{M} \in \instset\setminus\{M\}: \exists \ z \in \cone, \ \langle \mathrm{vect}\left({z(\pi-\pi^{\ast})^{\top}}\right),\mathrm{vect}\big({\tilde{M}}\big)\rangle=0 \right\}.$
\end{theorem}

\textit{Proof Intuition.}
First, we observe that an instance $\tilde{M}$ is in alternating set if there exists a $\pi \in \Pi \setminus \{\pi^{\ast}\}$ and $z \in \cC$, such that $z^{\top}\tilde{M}(\pi-\pi^*) > 0$. If $\pi$ and $\pi^*$ were pure strategies, it would have been exactly $\inf_{z\in\cC\setminus\{0\}} z^{\top}(\tilde{M}_a - \tilde{M}_{a^*}) > 0$. Let us denote the $z$ achieving $\inf$ as $z_{\inf}$, i.e., the preference for which $\tilde{M}_a$ and $\tilde{M}_{a^*}$ are the least distinguishable. Thus, we observe that $z^{\top}_{\inf} \tilde{M}_a$ functions exactly as the mean of the arm $a$ for an instance $\Tilde{M}$, while $z^{\top}_{\inf}(\tilde{M}_{a^*} - {M}_a)$ acts as the suboptimality gap of arm $a$. Now, we extend this idea in the classical lower bound scheme to get a nested optimization problem with $\inf$ over $z \in \cC$ and $\tilde{M}$ in the alternating set, and a sup over allocations $w \in \Delta_K$. We further show that the $\inf$ for $\tilde{M}$ appears at the boundary of the alternating set defined as $\partial\Lambda(M)$.  

\textbf{Discussions.} (i) \underline{Novelty:} In the best of our knowledge, this is the first lower bound for PrePEx with fixed confidence with an explicit KL-based dependence. All the existing lower bounds are gap dependent, and valid for a narrow range on mean vectors or known preference cone, i.e. the right orthant. Our proof does not need such assumptions. The gap-dependent bounds are special cases of ours (cf. Theorem~\ref{thm:gaussian-bandit} for Gaussian rewards).

(ii) \underline{Geometric Insights.} Theorem~\ref{thm:lower-bound} provides multiple geometric insights into the effect of the ordering cone $\cC$ on the characteristic time. \textit{First}, the alternating set $\altset{M}$ is piece-wise polyhedral and non-convex. We address consequent issues in Section~\ref{sec:relax-lower-bound}. \textit{Second}, there is an additional minimization over the vectors lying in the cone $\cC$. We interpret the minimization over vectors in the cone as a \textit{instance- and preference-dependent scalarization of the distance between the given instance $M$ and the corresponding most-confusing instance in $\altset{M}$}
. \textit{Third}, in the proof, we show that the reward gap using the best policy $\pi^*$ and a given policy $\pi$ for the most confusing instance belongs to the polar cone $\cC^{\circ}$ of the preference cone $\cC$. The most confusing lies on the boundary of this polar cone and its projection the policy gaps ($\pi^*-\pi$). Further insights can be obtained by imagining
the polar cone to be orthogonal to the cone $\cC$. Then, the vector of reward-gaps for the most confusing instance for every objective is orthogonal to the generating rays of $\cC$.
These novel geometric insights are complementary to the existing algebraic and statistical insights available in the lower bound literature~\citep{kone2023adaptive,ararat2023vector}.

\vspace*{-.5em}\subsection{Characteraization of Lower Bounds for Gaussians}\vspace*{-.5em}
To understand our lower bound better and to compare it with the literature, we present a reduction for Gaussian bandits. In Gaussian bandits, we assume that the reward vectors of arm $a \in [K]$ are generated from a multivariate Gaussian distribution $\cN(\mu_a, \Sigma)$, where the covariance is a diagonal matrix: $\Sigma \eqdef \mathrm{Diag}(\sigma_1^2, \ldots, \sigma_L^2)$.

\begin{theorem}[Lower Bound for Gaussian Bandits]
\label{thm:gaussian-bandit}
\begin{enumerate}[leftmargin=*]
\item Given any $\pi^\star\in \Pi^*(M)$ and $N(\pi^*)$ being the set of neighbouring policies of $\pi^*$, the most confusing instance of $M$ belongs to the set
\begin{align*}
\hspace*{-1em}    \Big\lbrace \tilde{M} \in \mathcal{M}\setminus \{M\}: \tilde{M}_{k,\ell} &= M_{k,\ell} - \beta \frac{\sigma_\ell^{2}}{z^\ell}  \frac{(\pi^* -\pi)_k}{ w_k}, ~~\forall \pi \in N(\pi^*), z \in \cC\setminus\{0\}, k\in [K], \ell\in [L] \Big\rbrace \, ,
\end{align*}
where $\beta \triangleq \frac{z^{\top}M(\pi^*-\pi)}{ \mathrm{Tr}(\Sigma) \|\pi^*-\pi\|_{\mathrm{Diag}(1/w)}^2}$.
\item The inverse of characteristic time, i.e. $(\cT^{\mathrm{Gauss}}_{M,\cC})^{-1}$, for an instance $(M, \cC)$ is
\begin{align}\hspace*{-1.4em}
\inf_{\substack{\pi \in N(\pi^*)\\ \pi^* \in \Pi^*(M)}}\min_{z \in \cC\setminus\{0\}} \frac{(z^{\top}M(\pi^*-\pi))^2}{2\mathrm{Tr}(\Sigma)\|\pi^*-\pi\|_2^2}\,.\notag
\end{align}
\end{enumerate}
\end{theorem}

\textbf{Consequences.} \textit{First}, we observe an interesting phenomenon that a bilinear projection of mean matrix $M$ on the preferences and policy gaps operates as an extension of suboptimality gap in classical BAI. This is a reminiscent of the lower bound for pure exploration under known linear constraints as in \cite{carlsson2024pure} who show that the hardness of the problem depends only on the projection of the mean vector on the policy gap. In addition to similar projection structure, preferences introduce a novel bilinearity here. \textit{Second}, we show how the lower bound inflates with the covariance matrix for each objective. This shows the richness of our KL-divergence based lower bound as opposed to gap-based bounds which have difficulty accommodating variance related terms directly.

\textit{Connection to existing results.} Our result generalizes several existing lower bounds for BAI. 

1. \textit{BAI lower bound.} Our lower bound is able to recover that of~\cite{kaufmann2016complexity} for the standard BAI problem with fixed confidence. In the case of the standard BAI problem, the ordering cone is given by $\cC\eqdef\mR_{+}$ and therefore the minimization over $\cC$ in~\eqref{eqn:characteristic-time} becomes redundant. The definition of the alternating set is then given by the set of instances which have a different optimal arm than $\mu$ which is exactly the set considered in~\citep{kaufmann2016complexity}.

2. \textit{Pure exploration under known constraints.} Our lower bound is able to recover the lower bound of~\cite{carlsson2023pure} for the BAI problem with fixed confidence and linear constraints. This is the case with $L=1$ and the ordering cone being $\cC\eqdef \mR_{+}$ making the minimization over $z \in \cC$ in~\eqref{eqn:characteristic-time} redundant.
$\altset{M}$ becomes $\altset{M}=\{\tilde{\mu}: \max_{\pi \in \Pi} \tilde{\mu}^{\top}\pi \ge \mu^{\top} \pi^\ast\}$ and we retrieve exactly their Corollary 1. 

\vspace*{-.5em}\section{Algorithm Design: PreTS}\vspace*{-.5em}
In this section, we propose an algorithm that tracks the lower bound. However, this is not straightforward since the alternating set is non-convex. We first propose a convex relaxation for this set and then, design a Track and Stop style algorithm, called PreTS. 

\vspace*{-.5em}\subsection{Convex Relaxation of the Lower Bound}\label{sec:relax-lower-bound}\vspace*{-.5em}
One of the major differences regarding the structure of lower bounds compared to a standard BAI problem is that $\altset{M}$ is a piece-wise polyhedron, i.e., a union of hyperplanes. Each hyperplance corresponds to a policy $\pi \in \Pi \setminus \{\pi^{\ast}\}$. 
To make the optimization problem tractable and obtain a convex program, we relax $\altset{M}$ using its convex closure, denoted by $\cc{\altset{M}}$.  We note that the construction of such a convex relaxation for track-and-stop (when the lower bound problem is non-convex) has been done in the MDP setting~\citep{al2021adaptive}. We define $\cc{\altset{M}}$ in Theorem~\ref{thm:ccl-alt-set} by formulating it as a disjunctive program, which we can reformulate further as a linear program~\citep{balas1985disjunctive}.
\begin{theorem}
\label{thm:ccl-alt-set}
Let $\cF \eqdef \cup_{\Pi \setminus \pi^{\ast}} \left\{\tilde{M} \in \instset:\ \exists \ z \in \cone, \langle \mathrm{vect}\left({z(\pi-\pi^{\ast})^{\top}}\right),\mathrm{vect}\big({\tilde{M}}\big)\rangle=0\right\}$. 
Fix $z \in \cone$ such that $z = \sum_{i} \alpha_{i}v_{i}$. Then, we have $\cc{\cF} = \cI$, where $\cI$ is defined as
\begin{align}\label{eq:conv_hull}
\cI \eqdef \lbrace \tilde{M} \in \instset \mid 
\gamma_i^{\top} \mathrm{vect}(\tilde{M}) = 0,~~ \forall~~ \gamma_i = \alpha_{i} \text{vect}(v_{i}(\pi-\pi^{\ast})^{\top}) 
\text{ and } \alpha_i \in \mathbb{R}_{\geq 0}\rbrace\,.
\end{align}
\end{theorem}
Using the convex hull (Equation~\eqref{eq:conv_hull}), we quantify the optimal value for a given allocation $w$ as 
\begin{align*}
    \overline{\cV}_{\cC}(w,M) \eqdef \min_{\tilde{M} \in \cc{\partial\altset{M}}} \inf_{z \in \cC} \sum_{k=1}^{K} w_k \kl{ z^{\top}M}{z^{\top}\tilde{M}}\,.
\end{align*} 
The corresponding optimal allocation is 
\begin{align}\label{eq:conv_lb}
\overline{w}^{\ast}(M) &= \arg\max_{w \in \Delta^{K}} \inf_{\substack{\pi \in \Pi\setminus \pi^*\\ \pi^* \in \Pi^*(M)}} \min_{\tilde{M} \in \cc{\partial\altset{M}}} \inf_{z \in \cC} \sum_{k=1}^{K} w_k \kl{z^\top M}{z^{\top}\tilde{M}}\,.\vspace*{-.5em}
\end{align} 
Hereafter, we consider Equation~\eqref{eq:conv_lb} as the optimization problem to be tracked.
To compute $\overline{\cV}_{\cC}(w,M)$, we need access to the true instance $M$ which is not available to us. Our Track-and-Stop strategy is based on repeatedly sampling an arm to construct an estimate of $M$, i.e. $M_t$, and exploiting continuity properties of $\overline{\cV}_{\cC}(w,M)$ to show that $\overline{\cV}_{\cC}(w,M_t) \rightarrow \overline{\cV}_{\cC}(w,M)$ and the cumulative number of arm plays  $N_{t,k} \rightarrow w_{k}, \ w_{k} \in \overline{w}^\ast(M)$. These properties ensure that it makes sense to design a Track and Stop style algorithm for this problem.

\begin{theorem}[Analytical Properties]\label{thm:perturbation-properties}
For all $M\in \mR^{L \times K}$ and all preference cones $\cC$, we get \textit{1.} The mapping $(w,M) \rightarrow \overline{\cV}_{\cC}(w,M)$ is continuous. 
\textit{2.} The characteristic time mapping $M \rightarrow \overline{\chartime}$ is continuous. 
\textit{3.} The set valued function $M \rightarrow \overline{w}^\ast(M)$ is upper-hemicontinuous.
\textit{4.} The set $\overline{w}^\ast(M)$ is convex.
\end{theorem}
\textit{Discussion: Cost of Convexification.} For Gaussian bandits, as we can get the analytical form of the most confusing instance $\tilde{M}$ (Theorem~\ref{thm:gaussian-bandit}), we do not pay any extra cost of convexification. In the non-Gaussian settings, where we cannot find such analytical forms for the most confusing instances, the minimum value of the inner minimisation problem under the convex hull (Equation~\eqref{eq:conv_lb}) can go lower than the minimum value found in the original non-convex set of instances (Equation~\eqref{eqn:characteristic-time}). 
Thus, the characteristic time attained by solving the convex relaxation might be higher than that of the original lower bound. Hence, an algorithm solving the convex relaxation has a higher stopping time. However, convexification is essential for the computational feasibility of a lower bound-tracking algorithm for PrePEx. This computational-statistical trade-off will be interesting to study in the future.

\vspace*{-.5em}\subsection{Algorithm: Preference-based Track-and-Stop (PreTS)}\vspace*{-.5em}
\setlength{\textfloatsep}{8pt}
\begin{algorithm}[t!]
\caption{Preference-based Track-and-Stop (PreTS)}
\label{algo:vector-tns}
\begin{algorithmic}[1]
\STATE{\textbf{Input:} Confidence parameter $\delta$, preference cone $\cC$}
\IF{$\sup_{\pi^*_t} \min_{\tilde{M} \in \cc{\partial\Lambda_{\pi^*_t}(\mestm)}}\min_{z \in \cC}\sum_{k}N_{k,t} \kl{z^{\top}\hat{M}^{(k)}_t}{z^\top\tilde{M}^{(k)}} \ge \beta(t,\delta)$}
\STATE{Compute $w_t \leftarrow \arg\max_{w \in \Delta^{K}} \overline{\cV}_{\cone}(w,\hat{M}_t)$}
\STATE{Play $k_t \leftarrow \arg\min_{k \in [K]} \big\vert N_{k,t} - \sum_{s=1}^{t}w_{s} \big\vert$}
\STATE{Observe reward $r_t$}
\STATE{Construct an empirical estimator $\hat{M}_t$}
\ENDIF 
\STATE{Construct a Pareto Front $\hat{\cP}_{\tau}$ from empirical means $\hat{M}_{\tau}$}
\STATE{Return: $\hat{\cP}_{\tau}$}
\end{algorithmic} 
\end{algorithm}

We now construct a general recipe to design a PrePEx algorithm when we do not have access to the true instance $M$. The fundamental element of any such recipe is constructing an estimate of $M$. 
For a given set of observed rewards $\{R_t\}_{t=1}^T$, we obtain a column-wise empirical average of the observed rewards and use it as our estimator of $M$. Now, we elaborate the three key components of our PrePEx algorithm Preference-based Track and Stop (PreTS, Algorithm~\ref{algo:vector-tns}). 

1. \textbf{Sampling Rule:} For the sampling rule, we consider a Track-and-Stop strategy~\citep{garivier2016optimal}. It tracks the optimal proportion of arm sampling by plugging in the empirical estimates of means and empirical count $N_{k,t}$ in the convexified lower bound. This leads to an allocation policy with improved information acquisition.

2. \textbf{Stopping Rule:} Our ultimate stopping goal is to identify arms that are on the Pareto front. Based on this, we define the confidence set as: 
\begin{equation}
\label{eqn:confidence-set}
\confball \eqdef \left\{\tilde{M} \in \mathcal{M}: \ \min_{z \in \cC} \sum_{k} N_{k,t} d_{\textrm{KL}}(z^{\top} \hat{M}^{(k)}_{t},z^{\top}\tilde{M}^{(k)}) \le \beta(t,\delta) \right\}\,,
\end{equation}
where $\beta(t,\delta) \triangleq \sum_{a \in S} 3 \ln \left(1+\ln \left({N}_{k,t}\right)\right)+K\mathcal{T}\left(\frac{\ln \left(\frac{1}{\delta}\right)}{K}\right)$ and $\mathcal{T}$ is defined in Equation~\eqref{eq:defn_mathcalT}. Our first claim is to show that the true instance belongs to the confidence ellipsoid with high probability.  

\begin{lemma}[Confidence Ball]
\label{lem:conf-ball}
For any $t \in \mathbb{N}$ and $c(t,\delta)$ is defined in Equation~\eqref{eqn:confidence-set}, we have 
$\mP\left(M \notin \confball \right) \le \delta$. 
\end{lemma}
Thus, we can now formalise the corresponding Chernoff-stopping rule as
\begin{equation}
\label{eqn:chernoff-stopping}
\min_{\tilde{M} \in \cc{\partial\altset{\mestm}}}\min_{z \in \cC}\sum_{k}N_{k,t} \kl{z^{\top}\hat{M}^{(k)}_t}{z^\top\tilde{M}^{(k)}} \ge \beta(t,\delta)
\end{equation}
Given the estimates $\hat{M}_t$, the problem in Equation~\eqref{eqn:chernoff-stopping} can be solved efficiently. Next, we show that upon stopping with Equation~\eqref{eqn:chernoff-stopping}, PreTS returns the true Pareto Front $\pftrue$ with probability $1-\delta$. Let $\pfestm$ denote the estimated Pareto Front at time $t$, which is constructed using estimates $\hat{M}_t$. Then at stopping time $\tau$, we have 
\begin{eqnarray*}
\mP\left( \pftrue \neq \pfestm \right) &\le& \mP\left( \exists \ t \in \mN: \sum_{k,\ell} N_{k,t} \kl{z^\top\hat{M}^{(k)}_t}{z^\top M^{(k)}} \ge \beta(t,\delta)  \right)\\
&\le& \sum_{t=1}^{\infty} \mP\left( \sum_{k,\ell} N_{k,t}\kl{z^\top\hat{M}^{(k)}_t}{z^\top M^{(k)}} \ge \beta(t,\delta) \right) \leq \delta
\end{eqnarray*}
where, the last inequality is true due to Theorem~\ref{thm:stopping-time}, a concentration result on the KL-divergence with preference projected mean rewards.
\begin{theorem}\label{thm:stopping-time}
For all $\rho \ge (K+1)$, $z \in \cC$ and $t \in \mN$, we have that:
\[
\mP\left[ \sum_{k\in[K]} N_{k,t} \kl{z^\top \hat{M}^{(k)}_t}{z^\top M^{(k)}} \ge \rho \right] \le \exp\left(-\rho\right)\left(\frac{\lceil \rho \log(t) \rceil}{K}\right)^{K} \exp\left(K+1\right) 
\]
\end{theorem}

3. \textbf{Recommendation Rule:} At the end of stopping time $\tau$, the algorithm returns an estimate of the Pareto Front $\hat{\cP}_{\tau}$. 

\vspace*{-.5em}\section{Upper Bound on Sample Complexity}\vspace*{-.5em}
Now, we prove upper bound on the expected sample complexity of PreTS. This requires us to the Track-an-Stop proof technique. But the challenge is to show concentration of the pareto fronts under a suitable metric.

\textbf{Concentrating to the Pareto Front.} 
To show that upon stopping the algorithm returns the true Pareto Frontier, we need to establish a valid metric to show such convergence.
Usually, the distance between sets is measured using the Hausdorff metric~\citep{costantini1995infimum}, i.e. $d_{H}( \pfestmstop,\cP) \eqdef \max\left\{ \sup_{k \in \pfestmstop} \inf_{k' \in \cP} \Vert M_{k}-M_{k'}\Vert_{\infty}, \ \sup_{k \in \cP} \inf_{k' \in \pfestmstop} \Vert M_{k}-M_{k'} \Vert_{\infty} \right\}$. 
But the Hausdorff distance only defines a pseudo-distance between sets and $\paretospace$ may not be closed under this metric. To circumvent this issue, we build upon the notion of a gap-based metric considered in the antecedent literature~\citep{auer2016pareto} to measure the distance between the mean reward of an arm and a given Pareto Front. We extend it to a distance metric between elements in the space of Pareto Fronts $\paretospace$.

\begin{definition}[Distance from Pareto Front]
The distance of the mean of arm $k$ from the Pareto Front $\pftrue$ is $d(k,\pftrue) \eqdef \inf_{\varepsilon \ge 0} \varepsilon$, such that  $M_{k}+\varepsilon\ones \vectneq  M_{k'},  \ k' \in \pftrue$. Equivalently,
\begin{align}
  d(k,\pftrue) &\triangleq \inf_{k' \in \pftrue}\max\left\{0,\sup_{z \in
                \cone\cap \mB(1)} z^{\top}\left(M_{k'}-M_{k}\right)\right\}. \label{eq:Delta-def}
\end{align}
\end{definition}

\begin{definition}[Distance between Pareto Fronts]
\label{defn:pareto-dist}
We define the metric between Pareto Fronts $\pdist{\cdot}{\cdot}: \paretospace \times \paretospace \rightarrow \mR_{\ge 0}$ as $
\pdist{\hat{P}}{\pftrue} \eqdef \max\left\{\sup_{k \in \hat{P}} d(k,\pftrue), \sup_{k \in \pftrue} d(k,\hat{P})\right\}$.
\end{definition}
In the appendix, we establish that (i) $d(\cdot,\cdot)$ is a valid metric on $\paretospace$, and (ii) $\paretospace$ is compact and complete under $d(\cdot,\cdot)$. Now, we leverage this metric to show that the Pareto Front defined by the arm-wise constructed estimator $\hat{M}_t$ concentrates towards the true Pareto Front.

\begin{theorem}[Concentration of mean estimates]
\label{thm:mean-concentration}
For any pair $(i,j) \in \armset \times \armset$ and $z \in \cone$, we have
\begin{align*}
\label{eqn:pairwise-concentration}
\Big\vert z^\top \left(M_i-M_j\right) - z^\top\left(\hat{M}_{i,t}-\hat{M}_{j,t}\right)\Big\vert \le \beta_{ij}(t)\,,
\end{align*}
where $\beta^2_{ij}(t) \eqdef 4\|z\|_1^2{\left(h\left(\frac{\log(\frac{K_1}{\delta})}{2}\right)+\sum_{a \in \{i,j\}} \log\left(4+\log(N_{a}(t))\right) \right) \left(\sum_{
a \in \{i,j\}}\frac{1}{N_{a}(t)}\right)}$,  $K_1 \eqdef \frac{K(K-1)}{2}$, and $h(x) \approx x+\log(x)$.
\end{theorem}
\textit{Proof Sketch.} This is a consequence of jointly applying a vectorial concentration result for multiple objectives of each arm~\citep{kaufmann2021mixture}, and pairwise time-uniform concentration bounds~\citep{kone2023adaptive}. A key observation here is that the confidence radii depends on the magnitude of the preference vector $z$ and scales with different objectives accordingly.

\textbf{Sample Complexity of PreTS.} Using this new concentration result for the Pareto Front and the stopping rule in Equation~\eqref{eqn:chernoff-stopping}, we derive an upper bound on the expected stopping time of PreTS.
\begin{theorem}[Upper Bound on Sample Complexity]
\label{thm:upper-tns}
For any $\alpha > 0$ and $c(t,\delta)$ defined in~\eqref{eqn:chernoff-stopping}, we have that the stopping time satisfies\vspace*{-.5em}
\[
\lim_{\delta \rightarrow 0} \frac{\mE[\stoptime]}{\log\left(\frac{1}{\delta}\right)} \le \overline{\chartime} \ \forall \ M \in \mR^{L \times K}
\]
\end{theorem}\vspace*{-.5em}
The basic outline of the proof follows a general strategy to prove Track-and-Stop result. However, the new arguments lie in establishing that the Pareto fronts converge under a suitable metric sufficiently fast. Our proof implies that PreTS matches the convex relaxation of the lower bound asymptotically at the corresponding risk level $\delta$. Strictly, speaking this is not \textit{asymptotically optimal} since, we do not track the exact lower-bound.

\vspace*{-.5em}\section{Conclusion and Future Works}\vspace*{-.5em}
We study the fixed-confidence version of preference-based pure exploration problem under linear stochastic bandit feedback, where each arm corresponds to a reward vector ordered according to a preference cone. 
We derive a novel lower bound for this problem. We leverage the lower bound further to derive a track-and-stop based algorithm for PrePEx problem. As future work, it would be interesting to verify our results on a real-world datasets. 

Additionally, it would be interesting and challenging to study how other asymptotically optimal pure exploration strategies, e.g. gamified explorers~\citep{Degenne2019PureEW}, top-two algorithms~\citep{jourdan2022top}, can be adapted to this setting. In general, improving the computational efficiency and studying the optimality gap with respect to the non-convex lower-bound problem would be of fundamental interest. 

\bibliographystyle{apalike}
\bibliography{shared/ref}

\newpage
\appendix
\part{Appendix}
\parttoc
\newpage
\section{Notations}\label{app:notations}
\renewcommand{\arraystretch}{1.5}
\begin{longtable}{l|l}
\hline
Notation & Description  \\
\hline  
 $\cC,\vectdom$ & Given convex cone and induced partial order \\
 \hline 
 $K,L$ & number of arms and objectives \\
 \hline
 $\pftrue,\pfestm$ & Ground truth Pareto set and estimated Pareto set \\
 \hline
 $\paretospace$ & Space of all Pareto Frontiers on $[0,1]^{K}$ \\
 \hline
 $M \in \mR^{K \times L}$ & matrix with mean reward of $K$ arms \\
 \hline
 $r_{k_t},M_{k_t},\eta_t$ & observed reward, mean reward and noise \\
 \hline
 $c(t,\delta)$ & Confidence ball at time $t$ with confidence $\delta$  \\
 \hline 
 $\hdist{X}{Y}$ & Hausdroff distance between sets $X$ and $Y$ \\
 \hline 
 $\pdist{P}{\hat{P}}$ & Distance metric between Pareto Fronts $P$ and $\hat{P}$ \\
 \hline
$w$ & Allocation vector \\
 \hline
 $\Pi$ & Family of policies \\
 \hline
$\muestm{k}{t},\mutrue{k}$ & Estimated and true of mean rewards \\
\hline
$\cc{S}$ & Convex hull of set $S$ \\
\hline
$\altset{M}$ & Set of alternating instances associated with $M$ \\
\hline
\caption{Table of Notations}\label{tab:Notation}
\end{longtable}

\clearpage\section{Proofs of Lower Bounds}
\subsection{Generic Lower Bound: Proof of Theorem~\ref{thm:lower-bound}}\label{app:them1_proof}
\begin{proof}
The proof follows the basic structure of constructing a lower bound as in~\cite{kaufmann2016complexity}. Recall that their inverse of characteristic time is given by
\begin{equation}
\label{eqn:kaufman-bound}
\cT_{} = \sup_{w \in \Pi}\inf_{\tilde{M} \in \altset{M}} 
\sum_{k} w_k \kl{M_k}{\tilde{M_k}}
\end{equation}
The main challenge for our setting is describing the alternating set $\altset{M}$ and scalarization of the given instance $M \in \instset$. Given a matrix of arm-objective mean-rewards $M$, ordering cone $\cC$ and a family of policies $\Pi$, the Pareto front is the set of optimal values (ordered wrt $\cone$) of 
\begin{equation}\label{eqn:vect-opt}
\max_{\pi \in \Pi} \ M \pi \,.
\end{equation}
 
\begin{itemize}[leftmargin=*]
\item \textbf{Step 1: Constructing set of alternative instances} Let $\pi^{\ast} \in \arg\max_{\pi \in \Pi} M\pi$.

The set of confusing instances given $\Pi$ and $\cone$ is the set of all matrices $\tilde{M}$ which have a different Pareto front than $M$ when using the policy $\pi^{\ast}$. Therefore, the optimal values of~\eqref{eqn:vect-opt} with instance $M$ are not-dominated by those of instance $\tilde{M}$. Hence, the set of alternative instances, $\altset{M}$ is given by:
\[
\altset{M} := \left\{\tilde{M} \in \instset: \max_{\pi \in \Pi} \tilde{M} \pi \vectneq \tilde{M} \pi^{\ast} \right\}
\]
Let $\pi^{\prime} \in \arg\max_{\pi \in \Pi} \tilde{M} \pi \ \text{over} \  \cC$ which implies $\tilde{M} \pi^{\prime} \vectneq \tilde{M} \pi^\ast$ or equivalently: 
\[
\exists \ z \in \cC, \ \pi \in \Pi\setminus\{\pi^\ast\} \ \text{s.t.} \ z^\top \tilde{M} \pi > z^\top \tilde{M} \pi^\ast
\]
Therefore, the alternative set can be written as:
\begin{align*}
\altset{M} &\eqdef \cup_{\pi \in \Pi\setminus\{\pi^\ast\}} \left\{ \tilde{M} \in \instset: \ \exists \ z \in \cC, z^\top \tilde{M}\pi > z^\top \tilde{M}\pi^{\ast} \right\} 
\\
&= \cup_{\pi \in \Pi\setminus\{\pi^\ast\}} \left\{\tilde{M} \in \instset: \exists \ z \in \cC: z^\top \tilde{M}\cdot\left(\pi-\pi^{\ast}\right) > 0 \right
\}
\end{align*}
where, $\cdot$ represents a bilinear product and, its complement is given by:
\begin{align*}
\overline{\altset{M}} &\eqdef \cap_{\pi \in \Pi \setminus\{\pi^\ast\}} \left\{\tilde{M} \in \instset: \forall \ z \in \cC: z^\top \tilde{M}\cdot\left(\pi-\pi^{\ast}\right) \le 0\right\} \\
&= \cap_{\pi \in \Pi\setminus \{\pi^\ast\}} \left\{\tilde{M} \in \instset: \tilde{M}\cdot\left(\pi-\pi^{\ast}\right) \in \cone^{\circ}\right\} 
\end{align*}
where, $\text{ri}(\cone^{\circ})$ denotes the relative interior of the polar cone to $\cone$. Since $\cone$ is a polyhederal cone, it is closed and convex and therefore, its polar cone is non-empty, closed and convex.  Therefore, $\overline{\altset{M}}$ is non-empty. 
\item \textbf{Step 2: Hardest instance lies on the boundary} We now show that given $\pi^\ast, \ \text{and}\  \pi$, the hardest instances $\tilde{M} \in \instset$ are such that: 
\[
\tilde{M}\cdot\left(\pi-\pi^{\ast}\right) \in \text{bd}(C^{\circ}).
\]

Fix $\pi^{\prime} \in \Pi \setminus \{\pi^{\ast}\}$ and let $M' \in \left\{\tilde{M} \in \instset: \tilde{M}\cdot(\pi^{\prime}-\pi^{\ast}) \in \text{ri}(\cone^{\circ})\right\}$. Then, by convexity of $\left\{\tilde{M} \in \instset: \exists \ z \in \cC: z^\top \tilde{M}\cdot\left(\pi-\pi^\ast\right) > 0 \right\}$ there exists $M'' \in \left\{\tilde{M} \in \instset: \tilde{M}\cdot(\pi^{\prime}-\pi^{\ast}) \in \text{bd}(\cone^{\circ})\right\}$ such that 
$\big\vert z^{\top}M_{k} - z^{\top} M^{\prime}_{k} \big\vert \ge \big\vert z^{\top} M_{k} - z^{\top} M^{\prime \prime}_{k} \big\vert , \forall \ z \in \cone$. Since $\kl{z^{\top}M}{\cdot}$ is decreasing, we have: $\kl{z^{\top}M_{k}}{z^{\top}M^{\prime}_{k}} \ge \kl{z^{\top}M_{k}}{z^{\top}M^{\prime\prime}_{k}}$. 

Using the above arguments we see that the minimum argument of $\sum_{k}w_k\kl{z^\top M^{\prime}}{z^\top M^{\prime \prime}}$ is such that $M'' \in \left\{\tilde{M} \in \instset: \tilde{M}\cdot(\pi-\pi^{\ast}) \in \text{bd}(\cone^{\circ})\right\}$. 

The optimization problem now becomes: 
\begin{align*}
&\sup_{w \in \Pi}\inf_{\tilde{M} \in \altset{M}} \inf_{z \in \cC} \sum_{k} w_k \kl{z^{\top}M_k}{z^{\top}\tilde{M_k}} \\
= &\sup_{w \in \Delta^{K}} \inf_{\substack{\pi \in \Pi\setminus \pi^*}} \min_{\tilde{M} \in \partial \altset{M}}\inf_{z \in \cC} \sum_{k=1}^{K} w_k\kl{z^{\top}M_{k}}{z^{\top}\tilde{M}_{k}}\,,
\end{align*}
where
\begin{align}
    \partial\altset{M} &\eqdef \cup_{\Pi \setminus \pi^{\ast}} \left\{\tilde{M} \in \instset: \exists \ z \in \cone, \ z^{\top}\tilde{M} (\pi-\pi^{\ast})=0 \right\} \notag\\
    &= \cup_{\Pi \setminus \pi^{\ast}} \left\{\tilde{M} \in \instset: \ \exists \ z \in \cone, \  \langle \text{vect}\left({z^{\top}(\pi-\pi^{\ast})}\right),\text{vect}({\tilde{M}})\rangle=0 \right\}\,.
\end{align}
\end{itemize}

Finally, to accommodate for multiple optimal policies $\pi^{\ast}$, the inner minimisation problem becomes 
\[
\sup_{w \in \Delta^{K}} \inf_{\substack{\pi \in \Pi\setminus \pi^*\\ \pi^* \in \Pi^*(M)}} \min_{\tilde{M} \in \partial \altset{M}} \inf_{z \in \cC} \sum_{k=1}^{K} w_k\kl{z^{\top}M_{k}}{z^{\top}\tilde{M}_{k}} 
\]
This concludes the proof.

\end{proof}

\subsection{Lower Bound for Gaussians: Proof of Theorem~\ref{thm:gaussian-bandit}}\label{sec:lb_gaussians}
\begin{proof}\,\\
\textbf{Step 1.} Simplifying the KL-divergence for a Gaussian bandit instance with identical variance across all objectives yields
\begin{equation}
\label{eqn:ctime}
\cV_{\cC}(w,M) = \min_{\tilde{M}\in \partial\altset{M}} \min_{z \in C }\sum_{k=1}^{K} w_k \sum_{\ell=1}^{L} (z^{(\ell)})^2 \frac{\left(\mu^{(\ell)}_k - \tilde{M}^{(\ell)}_k\right)^2}{2\sigma_\ell^2}\,.
\end{equation}
Recalling that due to the projection lemma, the $\tilde{M}$ achieving the minimum satisfies
\begin{equation}
\label{eqn:alt-set}
z^\top\sum_{k=1}^{K}\tilde{M}^\top_k\left(\pi^\ast_k - \pi_k\right) = 0, \ \forall z \in \cC
\end{equation}
Now, we formulate the Lagrangian of~\eqref{eqn:ctime} with dual variables $\beta$ for~\eqref{eqn:alt-set} as
\begin{align}
&\quad \cL\left(w, M, \tilde{M},\gamma,\beta\right)\notag\\
&= \sum_{k=1}^{K}  
w_{k} \sum_{\ell=1}^{L} (z^{(\ell)})^2 \left(\frac{\mu^{(\ell)}_{k} - \tilde{M}^{(\ell)}_{k}}{\sqrt{2}\sigma_\ell}\right)^{2} + \beta z^\top \tilde{M} \left(\pi^{\ast}-\pi\right) \nonumber \\
&= \sum_{k=1}^{K}  
w_{k} \sum_{\ell=1}^{L}  (z^{(\ell)})^2\left(\frac{\mu^{(\ell)}_{k} - \tilde{M}^{(\ell)}_{k}}{\sqrt{2}\sigma_\ell}\right)^{2} + \beta z^\top \tilde{M} \left(\pi^{\ast}-\pi\right)\label{eqn:lagrangian} 
\end{align}

\textbf{Step 2.} Taking the derivative w.r.t. $\tilde{M}$, we have
\begin{align*}
\frac{\partial \cL}{\partial \tilde{M}^{(\ell)}_{k}} &= w_{k} \left(z^{(\ell)}\right)^2 \left(\frac{-2}{2\sigma_\ell^{2}}\right) \left(\mu^{(\ell)}_{k} - {\tilde{M}}^{(\ell)}_{k}\right) + \beta z^{(\ell)}\left(\pi^{\ast}-\pi\right)_{k}\,,
\end{align*}
and setting it to zero, we get 
\begin{align}
&w_{k} \left(z^{(\ell)} \right)^2 \left(\frac{2}{2\sigma
_\ell^{2}}\right) \left(\mu^{(\ell)}_{k}-\tilde{M}^{(\ell)}_{k}\right) = \beta z^{(\ell)} \left(\pi^{\ast}-\pi\right)_{k} \nonumber \\
\implies &\tilde{M}^{(\ell)}_{k} = \mu^{(\ell)}_{k} - \frac{2\sigma_\ell^{2}\beta\left(z^{(\ell)}\left(\pi^{\ast}-\pi\right)_{k}\right)}{2w_{k}\left(z^{(\ell)}\right)^2} =\mu^{(\ell)}_{k}-\frac{\sigma_\ell^{2}\beta\left(\pi^{\ast}-\pi\right)_{k}}{w_{k} z^{(\ell)}} \,.\label{eqn:tilde-mu}
\end{align}
The last equality holds for any $z^{(\ell)}\neq 0$. Therefore, the Lagrangian now becomes 
\begin{align*}
&~~~~\cL(w, M, \beta)\\ 
&= \sum_{k=1}^{K} w_{k} \sum_{\ell=1}^{L} (z^{(\ell)})^2 \left(\frac{\mu^{(\ell)}_{k}- \tilde{M}^{(\ell)}_{k}}{\sqrt{2}\sigma_\ell}\right)^{2} + \beta \sum_{i=1}^{N} \lambda_{i} \sum_{\ell=1}^{L} v^{(\ell)}_{i} \sum_{k=1}^{K} \tilde{M}^{(\ell)}_{k} \left(\pi^{\ast}-\pi\right)_{k} \\
&= \sum_{k=1}^{K} w_{k} \sum_{\ell=1}^{L} (z^{(\ell)})^2 \frac{\left(\pi^{\ast}-\pi\right)^{2}_{k}}{(z^{(\ell)})^2 w^{2}_k}\frac{\beta^{2}\sigma_{\ell}^{2}}{2}+\beta \sum_{\ell=1}^{L}  z^{(\ell)} \sum_{k=1}^{K} \left(\pi^{\ast}-\pi\right)_{k} \left(\mu^{(\ell)}_{k} - \beta\sigma_\ell^{2} \frac{\left(\pi^{\ast}-\pi\right)_{k}}{z^{(\ell)} w_{k}}\right)\\
&= -\frac{\beta^{2}}{2} \sum_{k=1}^{K}\sum_{\ell=1}^{L} \sigma_\ell^2 \frac{\left(\pi^{\ast}-\pi\right)^{2}_{k}}{w_{k}}   + \beta \sum_{\ell=1}^{L}\sum_{k=1}^{K} z^{(\ell)} \left(\pi^{\ast}-\pi\right)_{k} \mu^{(\ell)}_{k}
\end{align*}
\textbf{Step 3.} By taking the derivative with respect to $\beta$, we have
\begin{align*}
\frac{\partial \cL}{\partial \beta}
&= -\beta \sum_{k=1}^{K}\sum_{\ell=1}^{L} \sigma_\ell^2 \frac{\left(\pi^{\ast}-\pi\right)^{2}_{k}}{w_{k}}   +  \sum_{\ell=1}^{L}\sum_{k=1}^{K} z^{(\ell)} \left(\pi^{\ast}-\pi\right)_{k} \mu^{(\ell)}_{k} \,,
\end{align*}
and setting it to zero, leads to: 
\begin{equation}
\label{eqn:beta}
\beta = \frac{\sum_{\ell=1}^{L}\sum_{k=1}^{K}z^{(\ell)}\left(\pi^{\ast}-\pi\right)_{k}\mu^{(\ell)}_{k}}{\sum_{k=1}^{K}\sum_{\ell=1}^{L} \sigma_\ell^2\frac{\left(\pi^{\ast}-\pi\right)_{k}^2}{w_k}} = \frac{z^{\top}M\Delta}{ \sigma_{o}^2 \|\Delta\|_{\mathrm{Diag}(1/w)}^2}\,,
\end{equation}
where $\sigma_{o}^2\triangleq \sum_{\ell=1}^{L} \sigma_\ell^2$, and $\Delta \triangleq \pi^{\ast}-\pi$.

From~\eqref{eqn:tilde-mu}, excluding for the origin lying within the cone, we get: 
\begin{align*}
\tilde{M}^{(\ell)}_{k} &= \mu^{(\ell)}_{k} - \beta\sigma_\ell^{2} \frac{\left(\pi^{\ast}-\pi\right)_{k}}{z^\ell w_k} \\
&= \mu^{(\ell)}_{k} - \frac{\sigma_\ell^{2}}{z^\ell}  \frac{\Delta_k}{ w_k} \frac{z^{\top}M\Delta}{ \sigma_{o}^2 \|\Delta\|_{\mathrm{Diag}(1/w)}^2}
\end{align*}
Finally, the Lagrangian from~\eqref{eqn:lagrangian} leads to
\begin{align*}
    &~~~~\cL(w,M)\\
    &= -\frac{1}{2}\left(\frac{z^{\top}M\Delta}{ \sigma_{o}^2 \|\Delta\|_{\mathrm{Diag}(1/w)}^2}\right)^2 \sigma_{o}^2 \|\Delta\|_{\mathrm{Diag}(1/w)}^2 +  \frac{ (z^{\top}M\Delta)^2}{ \sigma_{o}^2 \|\Delta\|_{\mathrm{Diag}(1/w)}^2}\\
    &= \frac{(z^{\top}M\Delta)^2}{2 \sigma_{o}^2 \|\Delta\|_{\mathrm{Diag}(1/w)}^2}\,.
\end{align*}

Thus, the characteristic time inverse is given by,
\begin{align*}
    &\max_{w \in \Delta^K}\inf_{\substack{\pi \in N(\pi^*)\\ \pi^* \in \Pi^*(M)}}\min_{z \in \cC\setminus\{0\}} \frac{(z^{\top}M\Delta)^2}{2\sigma_{o}^2 \|\Delta\|_{\mathrm{Diag}(1/w)}^2} = \inf_{\substack{\pi \in N(\pi^*)\\ \pi^* \in \Pi^*(M)}}\min_{z \in \cC\setminus\{0\}} \frac{(z^{\top}M\Delta)^2}{2\sigma_{o}^2 \|\Delta\|_2^2}\,.
\end{align*}
\end{proof}

\subsection{Proof of Theorem~\ref{thm:ccl-alt-set}}
This proof follows directly from Theorem~3.1 in~\cite{balas1985disjunctive}. Recall that the set $\cF$ is given by: 
\[
\cF \eqdef \cup_{\Pi \setminus \{\pi^{\ast}\}} \left\{  \tilde{M} \in \instset: \langle \text{vect}(z(\pi-\pi^{\ast})^{\top}), \text{vect}(\tilde{M}) \rangle =0 \right\}
\]
Any $z \in \cone$, we have $z = \sum_{i}\alpha_i v_i$. Rewriting, every hyperplane in $\cF$ as $P_{\pi} = \left\{ \tilde{M} \in \instset \big\vert \langle \text{vect}\left(\sum_{i} \alpha_{i} v_{i}(\pi-\pi^{*})^{\top}\right), \text{vect}(\tilde{M}) \rangle = 0\right\}$. Then, by Theorem~3.1 in~\cite{balas1985disjunctive}, $\cC(\cF)$ is given by: 
\begin{align*}
\cC(\cF) = \lbrace \tilde{M} \in \instset \mid 
\gamma_i^{\top} \text{vect}(\tilde{M}) \ge 0 \text{ and } \gamma_i^{\top} \text{vect}(\tilde{M}) \le 0,~~ \forall~~ \gamma_i = \alpha_{i} \text{vect}(v_{i}(\pi-\pi^{\ast})^{\top}) 
\text{ and } \alpha_i \in \mathbb{R}_{\geq 0}\rbrace\,.
\end{align*}

\subsection{Proof of Theorem~\ref{thm:perturbation-properties}}
Recall that: 
\begin{align*}
\bar{\cV}_{\cone}(w,M)&\eqdef \min_{\tilde{M}\in \cc{\partial\altset{M}}}\inf_{z \in \cone} \sum_{k=1}^{K} w_{k}\kl{z^{\top}M}{z^{\top}\tilde{M}} \\
\bar{w}^{\ast}(M) &\eqdef \arg\max_{w \in \Delta^{K}} \inf_{\substack{\pi \in \Pi\setminus \pi^*\\ \pi^* \in \Pi^*(M)}} \min_{\tilde{M}\in\cc{\partial\altset{M}}}\min_{z \in \cone} \sum_{k=1}^{K} w_{k}\kl{z^{\top}M}{z^{\top}\tilde{M}} 
\end{align*}
\begin{itemize}[leftmargin=*]
\item For $(1)$ and $(2)$ observe that $(z,M) \rightarrow z^{\top}M$ and $(z,M,\tilde{M})\rightarrow \kl{z^{\top}M}{z^{\top}\tilde{M}}$ are continuous maps for all $(z,M) \in \cone \times \instset$ and $(z,M,\tilde{M}) \in \cone \times \instset \times \cc{\partial\altset{M}}$. Further, $\sum_{k} w_{k} \kl{z^{\top}M}{z^{\top}\tilde{M}}$ is continuous in all its elements. Fix a sequence $(w_t,,z_t,M_t) \in \Pi \times \cone \times \instset$ such that $\left(w_t,z_t,M_t\right) \rightarrow \left(w,z,M\right)$. For any $\epsilon, \ \exists \ t' \ge 1$ such that $\Vert (w_t,z_t,M_t) - (w,z,M) \Vert \le \epsilon \ \forall \ t \ge t^{\prime}$.  Further, $\cc{\altset{M_t}} \rightarrow \cc{\altset{M}}$. Therefore, for every $\epsilon^{\prime}, \ \exists \ t^{\prime \prime} \ge 1$ such that $\forall \ t \ge t^{\prime \prime}$ we  
\[
\Big\vert \sum_{k} w_{k,t}\kl{z^{\top}_tM_t}{z^{\top}_t\tilde{M}_t}  - \sum_{k} w_{k}\kl{z^{\top}M}{z_t^{\top}\tilde{M}_t} \Big\vert \le \epsilon^{\prime} \ \forall \ \tilde{M}_t \in \mR^{K \times L}
\]
Taking $t \ge \max\{t^{\prime},t^{\prime \prime}\}$, we have:
\begin{align*}
& \Big\vert \inf_{\tilde{M} \in \partial\altset{M}}\inf_{z \in \cone} \sum_{k} w_{k,t}\kl{z^{\top}_tM_t}{z^{\top}_t\tilde{M}_t}  - \inf_{\tilde{M}\in \partial\altset{M}} \inf_{z \in \cone} \sum_{k} w_{k}\kl{z^{\top}M}{z_t^{\top}\tilde{M}_t} \Big\vert\\
&\le \Big\vert \inf_{\tilde{M} \in \partial\altset{M}}\inf_{z \in \cone} \sum_{k} w_{k,t}\kl{z^{\top}_tM_t}{z^{\top}_t\tilde{M}_t} - \sum_{k} w_{k}\kl{z^{\top}M}{z_t^{\top}\tilde{M}_t} \Big\vert \\
&\le \epsilon^{\prime}
\end{align*}

\item For $(3)$, we define $f(w,M) = \inf_{\tilde{M} \in \cc{\partial\altset{M}}} \inf_{z \in \cone}\sum_{k} w_k \kl{z^\top M}{z^\top \tilde{M}}$ and $C(w) = \Pi$. Then, from Berge's Theorem (Theorem~\ref{thm:berge} in Appendix), we get $w^{\ast}(M)$ is upper-hemicontinuous. 

\item For $(4)$, the convexity of $w^{\ast}(M)$ follows since the optimal solution 
\[
\max_{w \in \Pi} \inf_{\tilde{M} \in \cc{\partial\altset{M}}} \inf_{z \in \cone} \sum_{k} w_k  \kl{z^{\top}M}{z^{\top}\tilde{M}}
\] 
is concave for any given $\pi$ and $\pi^*$. 
\end{itemize}

\clearpage
\section{Proof of the Stopping Time}

\subsection{Proof of Lemma~\ref{lem:conf-ball}}
The proof follows by showing that $\inf_{z \in \cone}\sum_{k} N_{k,t} \kl{z^{\top}\hat{M}_{k,t}}{z^{\top}M_k}$ is an appropriate stochastic process and can be bounded using the mixture of martingales technique of~\cite{kaufmann2021mixture}. 

To this end, given a $z\in \cC$, we define the random variable as
\begin{align*}
    X_k(t) \triangleq \sum_{k} \max\{0,N_{k,t} \kl{z_{\inf}^{\top}\hat{M}_{k,t}}{z_{\inf}^{\top}M_k} - 3\ln(1+\ln N_{k,t})
    \}\,,
\end{align*}
where $z_{\inf} \triangleq \arg\inf_{z \in \cone}\sum_{k} N_{k,t} \kl{z^{\top}\hat{M}_{k,t}}{z^{\top}M_k}$

Since we assume $M_{k,\ell}$ belongs to exponential family of distributions, for any given $z_{\inf} \neq \boldsymbol{0}$, $z_{\inf}^{\top} M_{k}$ being a non-degenerate linear transform also belongs to the exponential family.

Now, plugging in $X_k(t)$ in Theorem 7 of \citep{kaufmann2021mixture} yields
\begin{align*}
&\mathbb{P}\left[\exists t \in \mathbb{N}: \sum_{k \in [K]} N_{k,t} \kl{z_{\inf}^{\top}\hat{M}_{k,t}}{z_{\inf}^{\top}M_k} \geq \sum_{k \in [K]} 3 \ln \left(1+\ln \left({N}_{k,t}\right)\right) + K \mathcal{T}\left(\frac{\ln \left(\frac{1}{\delta}\right)}{K}\right)\right]\\
\leq~ &\delta\,.    
\end{align*}

Here, $\mathcal{T}: \mathbb{R}^{+} \rightarrow \mathbb{R}^{+}$ is such that 
\begin{align}\label{eq:defn_mathcalT}
    \mathcal{T}(x) \triangleq 2 \tilde{h}_{3 / 2}\left(\frac{h^{-1}(1+x)+\ln \left(2\zeta(2)\right)}{2}\right)
\end{align} with
\begin{align}
\forall u \geq 1, \quad h(u) & =u-\ln (u) \\
\forall z \in[1, e], \forall x \geq 0, \quad \tilde{h}_z(x) & = \begin{cases}\exp \left(\frac{1}{h^{-1}(x)}\right) h^{-1}(x) & \text { if } x \geq h^{-1}\left(\frac{1}{\ln (z)}\right) \\
z(x-\ln (\ln (z))) & \text { else }\end{cases} \,,
\end{align}
and $\zeta(2) = \sum_{n=1}^{\infty} n^{-2}$.

\subsection{Proof of Theorem~\ref{thm:stopping-time}}
First, we prove the following lemma required to proceed with Theorem~\ref{thm:stopping-time}.
\begin{lemma}\label{lem:ind-concentration}
For any $k=1,2,\ldots,K$, let $1 \le t_{k} \le t$. Let $\eta >0$ and define the event: 
\[
C \triangleq \cap_{k \in [K]} C_{k} \triangleq \cap_{k \in [K]} \{t_k \le N_{k,t} \le (1+\eta) t_k\}
\] and let $\mathbbm{1}_{C}$ indicate whether the event holds. For $\rho \ge (1+\eta)K$, for all $z \in \cone$ we have:
\[
\mP\left[\ind{C}\sum_{k \in [K]} \numarmkt \klobjl \ge \rho \right] \le \left(\frac{\rho e}{K}\right)^{K}\exp\left(\frac{-\rho}{1+\eta}\right)\,.
\]
\end{lemma}
\begin{proof}
Fix $\zeta \in \mR^{K}_{+}$ and $t \ge 0$. Define $m_{k,t}$ such that:
\begin{eqnarray*}
m_{k,t} &=& 
\begin{cases}
m, \ \text{if} \ \exists \ 0 \le m \le z^{\top}M_{k}, \ \text{s.t.} \ t \kl{m}{z^{\top}M_{k}} = \zeta_{k} \\
0, \ \text{otherwise}
\end{cases}
\end{eqnarray*}
By monotonicity of $t\kl{}{}, \ t \rightarrow m_{k,t}$ is increasing. With $t=N_{k,t}$, we have that 
\[
N_{k,t}\kl{m_{k,N_{k,t}}}{z^{\top} M_{k}} = \zeta_{k} \le N_{k,t}\kl{z^{\top}\mklestm}{z^{\top}M_{k}}, \implies z^{\top} \mklestm \stackrel{(a)}{\le} m_{k,N_{k,t}} \stackrel{(b)}{\le} m_{k,(1+\eta)t_k},
\] 
where $(a)$ follows from monotonicity of $\kl{\cdot}{\cdot}$ and $(b)$ follows from monotonicity of $m^{(\ell)}_{k,t}$. 

With $t_k \kl{z^{\top}\hat{M}_{k,t_k(1+\eta)}}{z^{\top}M_{k}} = \frac{\zeta_{k}}{1+\eta}$ and non-negativity of $\kl{\cdot}{\cdot}$ we have:
\begin{eqnarray*}
&\quad& \mP\left( \cap_{k \in [K]}\left\{\mathbbm{1}_{C_k} N_{k,t} \klobjl \ge \zeta_{k}\right\} \right)\\
&\le& \mP\left( \cap_{k \in [K]} \left\{ z^{\top}\mklestm \le m_{k,N_{k,t}}, C_{k}\right\} \right) \\
&\le& \mP\left( \cap_{k \in [K]} \left\{ z^{\top}\mklestm \le m_{k,(1+\eta)t_{k}}, C_{k}\right\} \right) \\
&\le& \mP\left(\cap_{k \in [K]} \left\{ z^{\top}\mklestm \le m_{k,(1+\eta)t_{k}} , C_{k}\right\} \right) \\
&\stackrel{(a)}{\le}& \exp\left(-\sum_{k \in [K]}t_{k}\kl{m_{k,(1+\eta)t_{k}}}{z^{\top}M_{k}}\right)\\
&=& \exp\left(-\sum_{k \in [K]}\frac{\zeta_k}{1+\eta}\right)\,,
\end{eqnarray*}
where $(a)$ follows from Lemma~\ref{lem:kl-exp-bound}. 

Using Lemma~\ref{lem:multidim-exp-sum}, with $Z_{k} = \numarmkt \klobjl$ and $a = \frac{1}{(1+\eta)}$, we have that:
\[
\mP\left[\ind{C}\sum_{k \in [K]}\numarmkt \klobjl \ge \rho \right] \le \left(\frac{\rho e}{K}\right)^{K}\exp\left(\frac{-\rho}{1+\eta}\right)
\]
This concludes the proof.
\end{proof}

\textbf{Now, we provide the proof of Theorem~\ref{thm:stopping-time}.}
\begin{proof}
Let us define $D = \lceil \frac{\log(t)}{\log(1+\eta)} \rceil$  and set $\mathcal{D} = \{1,2,\ldots,D\}^{K}$. Let us define
\begin{align*}
A = \left\{\sum_{k \in [K]} N_{k,t} \klobjl \ge \rho \right\}
\\
B_d = \cap_{k=1}^{K} \left\{ (1+\xi)^{d_k-1} \le N_{k,t} \le (1+\xi)^{d_k} \right\} 
\end{align*}
We have $A = \cup_{d \in \cD} (A \cap B_d)$, hence $\mP(A) \le \sum_{d \in \cD} \mP(A \cap B_{d})$. 

For $\eta = \frac{1}{\rho-1}$ and $\rho \ge (1+\eta)K$, we get $\rho \ge K+1$.

Now, we use Lemma~\ref{lem:ind-concentration} with  $\eta = \frac{1}{\rho-1}$ and $\bar{t}_{k} = \left(1+\eta\right)^{d_k-1}$ to obtain for all $d \in \cD$:
\[
\mP\left( A \cap B_d \right) \le \left(\frac{\rho e}{K}\right)^{K} \exp\left(\frac{-\rho}{(1+\eta)}\right)
\]

By a union bound on $\cD$, we have: 
\[
\mP\left(A\right) \le  \left(\frac{D\rho e}{K}\right)^{K} \exp\left(\frac{-\rho}{1+\eta}\right) 
\]
Noting that $D = \lceil \frac{\log(t)}{\log(1+\eta)} \rceil$ and $\eta = \frac{1}{\rho-1}$, we get: 
\[
\mP\left(A\right) \le \exp\left(-\rho\right) \left(\frac{\rho \lceil \rho \log(t)\rceil}{K}\right)^{K} e^{K+1}\,.
\]

\end{proof}
\clearpage\section{Proofs for Sample Complexity Upper Bound}

\subsection{Pairwise Concentration Bounds}
\begin{lemma}[Pairwise concentration (Proof of Theorem~\ref{thm:mean-concentration})]
\label{lem:pairwise-concentration}
Consider the event: 
\begin{equation}
\label{def:conc-event}
\concevent_{t} \eqdef \cap_{i \in \armset} \cap_{j \neq i} \left\{ L_{i,j}(t) \le z^{\top}\mu_{i} - z^{\top}\mu_{j} \le  U_{i,j}(t) \right\}\,
\end{equation}
where $L_{ij}(t)=z^{\top}(\hat{\mu}_{i,t}-\hat{\mu}_{j,t})-\beta_{ij}(t)$ and $U_{ij}(t)=z^{\top}(\hat{\mu}_{i,t}-\hat{\mu}_{j,t})+\beta_{ij}(t)$, and $\beta_{ij}(t)$ is defined as
\begin{equation*}
    \beta^2_{ij}(t) \eqdef 4\|z\|_1^2{\left(h\left(\frac{\log(\frac{K_1}{\delta})}{2}\right)+\sum_{a \in \{i,j\}} \log\left(4+\log(N_{a}(t))\right) \right) \left(\sum_{
a \in \{i,j\}}\frac{1}{N_{a}(t)}\right)}\,.
\end{equation*}
Here, $K_1 \eqdef \frac{K(K-1)}{2},~ h(\cdot) \approx x+\log(1+x)$.
Then, we get 
\begin{equation*}
    \mP\left(\cap_{t=1}^{\infty} \cE_{t} \right) \ge 1-\delta\,.
\end{equation*} 
\end{lemma}

\begin{proof}
We have the following: 
\begin{eqnarray*}
\cE_t &=& \cap_{(i,j)} \left\{L_{i,j} \le z^{\top}\mu_{i} - z^{\top}\mu_{j}  \le U_{i,j} \right\} \\
&=& \cap_{(i,j)} \left\{ \big\vert z^{\top}\left(\hat{\mu}_{i,t}-\hat{\mu}_{j,t}\right) - z^{\top}\left(\mu_{i}-\mu_{j}\right) \big\vert \le \beta_{ij}(t)  \right\}
\end{eqnarray*}
where, $(i,j) \in \armset \times \armset$ is the set of arm pairs. By a union bound, we have the following: 
\begin{eqnarray*}
\mP\left(\overline{\cE}_t\right) &=& \mP\left(\exists \ t \ge 1: \overline{\cE}_{t} \ \text{ holds }\right)\\
&=& \mP\left(\exists \ t \ge 1: \big\vert z^{\top}\left(\hat{\mu}_{i,t}-\hat{\mu}_{j,t}\right) - z^{\top}\left(\mu_{i}-\mu_{j}\right) \big\vert \ge \beta_{ij}(t) \right)\\
&\le& \mP\left(\exists \ t \ge 1: \vert z^{\top}(\hat{\mu}_{i,t}-\mu_{i}) - z^{\top}(\hat{\mu}_{j,t}-\mu_{j}) \vert \ge \beta_{ij}(t)\right)\\
&\stackrel{(a)}{\le}& \sum_{(i,j)} \frac{\delta}{K(K-1)} \\
&=& \delta
\end{eqnarray*}
$(a)$ follows from 
\begin{enumerate}
    \item $z^{\top}(\hat{\mu}_{i,t}-\mu_{i})$ is a $\|z\|_1$ sub-Gaussian as $\hat{\mu}_{i,t}-\mu_{i}$ is 1-sub-Gaussian, and
    \item Lemma 7 of~\citep{kone2023adaptive} that states that for two 1-sub-Gaussian and centred random variables $X$ and $Y$, the following holds true with probability $1-\delta$.
    \[
    \left\vert \frac{1}{p}\sum_{i} X_i - \frac{1}{q}\sum_{i} Y_i \right\vert \le 2\sqrt{\left(h\left(\frac{\log(\frac{1}{\delta})}{2}\right) + \log \log(e^{4}p) + \log \log(e^{4}q)\right)\left(\frac{1}{p} + \frac{1}{q}\right)}\,,
    \]
    given $h(x) \approx x + \ln x$.
\end{enumerate}
\end{proof}

We now show that the Pareto fronts under the metric $d_{p}$.

\begin{lemma}
$(\cZ,d_p)$ is a complete metric space.
\end{lemma}

\begin{proof}
From Definition~\ref{defn:pareto-dist}, for two Pareto fronts $\cP_1,\cP_2 \in \cZ$, we have that:
\[
d_p(\cP_1,\cP_2) \eqdef \max\left\{\sup_{k \in \cP_1} d(k,\cP_1), \sup_{k \in \cP_2} d(k,\cP_1)\right\}
\]
where, 
\[
  d(k,\cP) = \inf_{k' \in \cP}\max\left\{0,\sup_{z \in
                \cone\cap \mB(1)} z^{\top}\left(\mu_{k'}-\mu_{k}\right)\right\}
\]

\begin{enumerate}[leftmargin=*]
\item We first show that $d_p(\cP_1,\cP_2)$ is a metric. Let $\cP_1, \cP_2 \in \cZ$. To show that $d_p$ is a metric, we show that: 
\begin{enumerate}
\item Symmetry: $d_{p}(\cP_1, \cP_2)$ is symmetric by definition
\item Triangle Inequality: We show that $d_{p}(\cP_1,\cP_3) \le 
d_p(\cP_1,\cP_2)+d_{p}(\cP_2,\cP_3)$.
\begin{eqnarray*}
d_{p}\left(\cP_1,\cP_3\right) &=& \max\left\{ \max_{k \in \cP_1} \min_{k' \in \cP_3} \mu_{k'}(X_{3}) -\mu_{k}(X_1) , \max_{k \in \cP_3} \min_{k' \in \cP_1} \mu_{k'}(X_1) - \mu_{k}(X_{3}) \right\} 
\end{eqnarray*}
We have that: 
\begin{eqnarray*}
&& \max_{k \in \cP_1} \min_{k' \in \cP_3} \mu_{k'}(X_{3}) -\mu_{k}(X_1) \\
&\le& \max_{k \in \cP_1}\min_{k' \in \cP_3} \mu_{k'}(X_3) + \min_{k'' \in \cP_2} \mu_{k''}(X_2) - \max_{k''\in \cP_2}\mu_{k''}(X_2) - \mu_{k}(X_1) \\
&\le& \max_{k'' \in \cP(X_2)} \min_{k' \in \cP_3} \mu_{k'}(X_3) -\mu_{k''}(X_2) + \max_{k \in \cP_1}  \min_{k'' \in \cP(X_2)} \mu_{k''}(X_2) - \mu_{k}(X_1)
\end{eqnarray*}
Using a similar argument: 
\begin{align*}
\max_{k \in \cP_3} \min_{k' \in \cP_1} \mu_{k'}(X_1) - \mu_{k}(X_{2}) \le \max_{k'' \in \cP(X_2)} \min_{k' \in \cP_1} \mu_{k'}(X_1) -\mu_{k''}(X_2) + \max_{k \in \cP_3}  \min_{k'' \in \cP(X_2)} \mu_{k''}(X_2) - \mu_{k}(X_3)
\end{align*}
Noting that for any positive numbers $a,b,c,d, \ \max\{a+b,c+d\} = \max\{a+c,b+d\}$, we have:
{\small
\begin{align*}
d_p(\cP_1,\cP_3)&\le& \max\Big\{ \max_{k'' \in \cP(X_2)} \min_{k' \in \cP_3} \mu_{k'}(X_3) -\mu_{k''}(X_2) + \max_{k \in \cP_1}  \min_{k'' \in \cP(X_2)} \mu_{k''}(X_2) - \mu_{k}(X_1), \\
&& \max_{k'' \in \cP(X_2)} \min_{k' \in \cP_1} \mu_{k'}(X_1) -\mu_{k''}(X_2) + \max_{k \in \cP_3}  \min_{k'' \in \cP(X_2)} \mu_{k''}(X_2) - \mu_{k}(X_3) \Big\} \\
&=& \max\Big\{\max_{k'' \in \cP(X_2)} \min_{k' \in \cP_1} \mu_{k'}(X_1) -\mu_{k''}(X_2) + \max_{k' \in \cP_1}  \min_{k'' \in \cP(X_2)} \mu_{k''}(X_2) - \mu_{k'}(X_1) , \\
&& \max_{k'' \in \cP(X_2)} \min_{k' \in \cP_3} \mu_{k'}(X_3) -\mu_{k''}(X_2) + \max_{k \in \cP_3}  \min_{k'' \in \cP(X_2)} \mu_{k''}(X_2) - \mu_{k}(X_3) \Big\} \\
&=& d_p(\cP_1,\cP_2) + d_p(\cP_2,\cP_3) 
\end{align*}
}

\item We now show that $d_{p}(\cP_1,\cP_{2}
) = 0 \iff \cP_{1} = \cP_{2}$. The implication $\cP_1 = \cP_{2} \implies d_{p}(\cP_1,\cP_{2})=0$ is immediate. For the other side, note that by Definition~\ref{defn:pareto-dist}, we have:
\begin{align*}
&& d_{p}\left(\cP_1,\cP_2\right) = 0  \\
&\implies& \sup_{k \in \cP_{1}} \Delta(k,\cP_{2})=0 \ \text{ and } 
\sup_{k \in \cP_{2}} \Delta(k,\cP_{1}) = 0 
\end{align*}
Further, $\sup_{k \in \cP_{1}} \Delta(k,\cP_{2})=0$ implies:
\begin{align*}
\forall \ k \in \cP_{1}, k \notdom k', \ k' \in \cP_{2} \iff \forall \ k \in \cP_1, \ k \in \cP_2
\end{align*}
A similar agrument using $\sup_{k \in \cP_{1}} \Delta(k,\cP_1)=0$ implies that $\forall \ k \in \cP_{2}, \ k \notdom k', k' \in \cP_1$. 
\end{enumerate}
\item We now show that $\pspace$ is compact under the metric $d_{p}$. Consider a sequence of Pareto fronts $\cP_1,\cP_2,\ldots,\cP_n \in \pspace$ and $\cP$ be the candidate for limiting Pareto front.
\begin{itemize}
\item Boundedness of $\cP$ is immediate. 

\item $\cP_n \rightarrow \cP, \ \text{therefore,} \ \forall \ \epsilon > 0, \exists \ N(\epsilon) \ \text{s.t.} \ \forall \ n > N(\epsilon)$ and $d_{p}(P_n,P) < \epsilon$. Let $\mu_{k}$ be a limit point of $\cP$, i.e., $\exists \ \text{a sequence} \ \mu_{k,n} \in \cP \ \text{such that} \ \mu_{k,n} \rightarrow \mu_{k}$. Since $d_{p}\left(\cP_n,\cP\right) \rightarrow 0$ for each $\mu_{k,n} \in \cP$ there exists $\mu_{k,n,m} \in \cP_n \ \text{s.t.} \ \mu_{k,n,m} \rightarrow \mu_{k,n}$. Using a diagonalization argument, we can obtain a subsequence $\mu_{k,n,m} \rightarrow \mu_{k}$. 
\item Since $\cP_n$ is compact, $\mu_k$ must lie in $\cP$ and therefore, $\cP$ is closed.
\end{itemize}
\end{enumerate}
\end{proof}

\subsection{Concentration to the Pareto Front}

\begin{lemma}
\label{lem:good-event}
There exists constants $C>0$ such that: 
\[
\mP\left( \bar{\cG}_{T} \right) \le 2 K^2 \frac{1}{1-\exp(-C)} T \exp\left(-C T^{1/8}\right)\,,
\]
where $\goodevent = \cap_{t = h(T)}^{T} \left\{ \pdist{\pfestm}{\pftrue} \le \epsilon \right\}$.
\end{lemma}

\begin{proof}
We then have that: 
\begin{eqnarray*}
\mP\left( \pdist{\pfestm}{\pftrue} \ge \epsilon \right) &=& \mP\left( \max \left\{ \pdist{\pfestm}{\pftrue}, \pdist{\pftrue}{\pfestm} \right\} \ge \epsilon \right) \\  
&\le& \mP\left(\pdist{\pfestm}{\pftrue} \ge \epsilon \right) + \mP\left(\pdist{\pftrue} {\pfestm} \ge \epsilon \right)
\end{eqnarray*}
Focusing on the first term we have: 
\begin{align*}
\pdist{\pfestm}{\pftrue} &= \inf_{k \in \pfestm}\sup_{k'\in\pftrue} \max\left\{0,\max_{z \in \cone \cap \mB(1)} z^{\top} \left(\mutrue{k'}-\muestm{k}{t}\right) \right\}\\
&\stackrel{(a)}{=} \min_{k \in \pfestm}\max_{k'\in\pftrue} \max\left\{0,\max_{z \in \cone \cap \mB(1)} z^{\top} \left(\mutrue{k'}-\muestm{k'}{t}+\muestm{k'}{t}-\muestm{k}{t}\right) \right\} \\
&\stackrel{(b)}{\le} \max_{k' \in \pftrue} \max\left\{0,\max_{z \in \cone \cap \mB(1)} z^{\top} \left(\mutrue{k'}-\muestm{k'}{t}\right) \right\} \\
&\quad+ \max_{k' \in \pftrue} \max\left\{0,\max_{z \in \cone \cap \mB(1)} z^{\top} \left(\muestm{k'}{t}-\muestm{k}{t}\right) \right\}
\end{align*}
Recall that $\bar{\cG}_T = \cup_{t=h(T)}^{T} \left\{\pdist{\pfestm}{\pftrue} \ge \epsilon \right\}$, then using a union bound, we have:
\begin{eqnarray*}
&~~& \mP\left(\bar{\cG}_{T} \right)\\ &\le& \sum_{t=h(T)}^{T} \mP\left( \pdist{\pfestm}{\pftrue} \ge \epsilon \right) \\
&\le& \sum_{t=h(T)}^{T} \Bigg[\sum_{(k,k')} \mP\left( z^{\top}\left(\muestm{k}{t}-\muestm{k'}{t}\right) \le z^{\top}\left(\mutrue{k}-\mutrue{k'}\right)-\xi \right)\\
&~~&\qquad\qquad +\mP\left( z^{\top}\left(\muestm{k}{t}-\muestm{k'}{t}\right) \ge z^{\top}\left(\mutrue{k}-\mutrue{k'}\right)+\xi \right)\Bigg] \,.
\end{eqnarray*}
Let $T$ be such that $h(T) \ge K^{2}$. Since $t\geq h(T)$, one has that $N_{k,t} \ge \sqrt{t}-K$. 

Then using a union bound on each pair and number of arms, as well as a Chernoff bound, we have that each of the terms in the above inequality are bounded as
\begin{eqnarray*}
&~~& \mP\left( z^{\top}\left(\muestm{k}{t}-\muestm{k'}{t}\right)
\le z^{\top}\left(\mutrue{k} - \mutrue{k'}\right) - \xi \right)\\
&=& \mP\left( z^{\top}\left(\muestm{k}{t}-\muestm{k'}{t}\right) \le z^{\top}\left(\mutrue{k} - \mutrue{k'}\right) - \xi, N_{k,t} \ge \sqrt{t} \right) \\
&\le& \sum_{s=\sqrt{t}-{K}}^{t} \mP\left( z^{\top}\left(\muestm{k}{s}-\muestm{k'}{s}\right) \le z^{\top}\left(\mutrue{k} - \mutrue{k'}\right) - \xi\right) \\
&\le& \sum_{s=\sqrt{t}-{K}}^{t} \mP\left( z^{\top}\left(\muestm{k}{s}-\mutrue{k}\right) - z^{\top}\left(\muestm{k'}{s} - \mutrue{k'}\right) \le - \xi/2 - \xi/2\right) \\
&\le& \sum_{s=\sqrt{t}-{K}}^{t} \left(\mP\left( z^{\top}\left(\muestm{k}{s}-\mutrue{k}\right) \le - \xi/2 \right) + \mP\left( z^{\top}\left(\muestm{k'}{s} - \mutrue{k'}\right) \ge \xi/2\right)\right) \\
&\le& \sum_{s=\sqrt{t}-{K}}^t \left( \exp\left( -s\kl{z^{\top}{M}_{k}-\xi/2}{z^{\top}M_{k}}\right) + \exp\left( -s\kl{z^{\top}{M}_{k'}+\xi/2}{z^{\top}M_{k'}}\right)\right) \\
&\le& \frac{\exp\left(-(\sqrt{t}-{K})\kl{z^{\top}{M}_{k}-\xi/2}{z^{\top}M_{k}}\right)}{1-\exp\left(-\kl{z^{\top}{M}_{k}-\xi/2}{z^{\top}M_{k}}\right)} + \frac{\exp\left(-(\sqrt{t}-{K})\kl{z^{\top}{M}_{k'}+\xi/2}{z^{\top}M_{k'}}\right)}{1-\exp\left(-\kl{z^{\top}{M}_{k'}+\xi/2}{z^{\top}M_{k'}}\right)}\\
&\leq& \left(\frac{1}{1-\exp\left(-\kl{z^{\top}{M}_{k}-\xi/2}{z^{\top}M_{k}}\right)} + \frac{1}{1-\exp\left(-\kl{z^{\top}{M}_{k'}+\xi/2}{z^{\top}M_{k'}}\right)}\right)\\
&~~&~~~~\times \exp\left(-(\sqrt{t}-{K})\min_{k,k'}\{ \kl{z^{\top}{M}_{k}-\xi/2}{z^{\top}M_{k}}, \kl{z^{\top}{M}_{k'}+\xi/2}{z^{\top}M_{k'}}\}\right)
\end{eqnarray*}
Similarly, we have: 
\begin{eqnarray*}
&~~& \mP\left(z^{\top}\left(\muestm{k}{t}-\muestm{k'}{t}\right) \ge z^{\top}\left(\mutrue{k}-\mutrue{k'}\right)+\xi\right)\\
&\leq& \left(\frac{1}{1-\exp\left(-\kl{z^{\top}{M}_{k}+\xi/2}{z^{\top}M_{k}}\right)} + \frac{1}{1-\exp\left(-\kl{z^{\top}{M}_{k'}-\xi/2}{z^{\top}M_{k'}}\right)}\right)\\
&~~&~~~~\times \exp\left(-(\sqrt{t}-{K})\min_{k,k'}\{ \kl{z^{\top}{M}_{k}+\xi/2}{z^{\top}M_{k}}, \kl{z^{\top}{M}_{k'}-\xi/2}{z^{\top}M_{k'}}\}\right)
\end{eqnarray*}
Now, we set 
$$C \triangleq \min_{k} \left\{\kl{z^{\top}M_k+\xi/2}{z^{\top}M_{k}},\kl{z^{\top}M_k-\xi/2}{z^{\top}M_{k}}\right\}$$ and 
\begin{align*}
    B \triangleq \sum_{(k,k')} &\Big(\frac{1}{1-\exp\left(-\kl{z^{\top}{M}_{k}-\xi/2}{z^{\top}M_{k}}\right)} + \frac{1}{1-\exp\left(-\kl{z^{\top}{M}_{k'}+\xi/2}{z^{\top}M_{k'}}\right)}\\
    &+ \frac{1}{1-\exp\left(-\kl{z^{\top}{M}_{k}+\xi/2}{z^{\top}M_{k}}\right)} + \frac{1}{1-\exp\left(-\kl{z^{\top}{M}_{k'}-\xi/2}{z^{\top}M_{k'}}\right)}\Big)\,.
\end{align*}
We observe that $B\leq 2 K^2 \frac{1}{1-\exp(-C)}$. Now, we get that
\[
\mP\left( \bar{\cG}_{T} \right) \le \sum_{t=h(T)}^{T} B\exp{-C\sqrt{t}} \leq 2 K^2 \frac{1}{1-\exp(-C)} T \exp\left(-C T^{1/8}\right)\,.
\]
\end{proof}

\subsection{Proof of Theorem~\ref{thm:upper-tns}}
\begin{theorem}[Restating Theorem~\ref{thm:upper-tns}]
For any $\alpha > 0$ and $c(t,\delta)$ defined in~\eqref{eqn:chernoff-stopping}, we have that the stopping time satisfies :
\[
\lim_{\delta \rightarrow 0} \frac{\mE[\stoptime]}{\log\left(\frac{1}{\delta}\right)} \le \alpha \bar{\cT}_{\cF}(M) \ \forall \ M \in \mR^{K \times L}
\]
\end{theorem}

\begin{proof}
\textbf{Step 1: Good Event}
Let $T \in \mN, \ \text{and } h(T) = \sqrt{T}$, define the good event: 
\begin{equation}
\label{eqn:good-event}
\goodevent = \cap_{t = h(T)}^{T} \left\{ d_p(\pfestm,\pftrue) \le f(\epsilon) \right\}
\end{equation}
where, $f(\epsilon)$ is such that: 
\[
d_p(\pfestm,\pftrue) \le f(\epsilon) \implies \sup_{w' \in w^{\ast}(\pfestmstop)} \sup_{w \in w^{\ast}(\pftrue)} \Vert w' - w \Vert \le \epsilon
\]

\textbf{Step 2: Concentration of Good Event}
In Lemma~\ref{lem:good-event}, we show that: 
\[
\mP\left(\badevent \right) \le 2 K^2 \frac{1}{1-\exp(-C)}T \exp\left(-C T^{\frac{1}{8}}\right)
\]
\textbf{Step 3: Tracking Lemma}
We have: 
\begin{eqnarray*}
\Big\vert N_{k,t} - w^{\ast}_{k}(M) \Big\vert &\le&  \Big\vert \frac{N_{k,t}}{t} - \frac{1}{t} \sum_{s=1}^{t-1} w^{\ast}_{k}(\hat{M}_s) \Big\vert + \Big\vert \frac{1}{t}\sum_{s=1}^{t-1} w^{\ast}_{k}(\hat{M}_{s}) - w^{\ast}_{k}(M) \Big\vert \\ 
&\le& \frac{N_{k,t}}{t} + \frac{h(T)}{t} + \Big\vert \frac{1}{t} \sum_{s=1}^{t-1} w^{\ast}_{k}(\hat{M}_{s}) - w^{\ast}_{k}(M) \Big\vert \\
&\le& \frac{K(\sqrt{t}+1)}{t} + \frac{h(T)}{t} + \Big\vert \frac{1}{t} \sum_{s=1}^{t} w^{\ast}_{k}(\hat{M}_{s}) - w^{\ast}_{k}(M) \Big\vert \\
\end{eqnarray*}

From~\citep{garivier2016optimal}, we have that:
\[
\max_{k } \Big\vert N_{k,t} - \sum_{t}
w_{k,t} \Big\vert \le K \left(1+\sqrt{t}\right) 
\]

\textbf{Step 4: Complexity of the good event}
\medskip 

Assume $t \ge T_{\epsilon}$, and let:
\[
C_{\epsilon}\left(M\right) \eqdef \inf_{w',M'} \overline{\cV}_{\cC}\left(w,M\right), \ \forall (w,M) \text{ s.t. } \ \Vert w' - w \Vert \le 3\epsilon, \ \pdist{\hat{P}_{t}}{\pftrue} \le  f(\epsilon)
\]
Then, we have that: 
\[
\overline{\cV}_{\cone}\left(N_{t},\hat{M}_t\right) \ge t C_{\epsilon}(M) 
\]

\textbf{Step 5: Bounding the stopping time for good and bad events}
\medskip

Let $\tau_{\delta}$ be the stopping time, then: 
\[
\min\{\tau_{\delta},T\} \le \sqrt{T} + \sum_{t=T_{\epsilon}}^{T} \ind{\tau_{\delta} \ge t}  
\]
From the stopping rule (Equation~\eqref{eqn:chernoff-stopping}), we get that: 
\begin{eqnarray*}
T_{\epsilon} + \sum_{t=T_{\epsilon}}^{T} \ind{ \overline{\cV}_{\cone}\left(N_t,\hat{M}_{t}\right) \le \beta(t,\delta)} &\le& \sqrt{T} + \sum_{t=T_{\epsilon}}^{T} \ind{ t C_{\epsilon}(M) \le \beta(t,\delta)} \\ 
&\le& \sqrt{T} + \frac{\beta(t,\delta)}{C_{\epsilon}(M)} 
\end{eqnarray*}
where, $\beta(t,\delta)$ is defined in Equation~\eqref{eqn:confidence-set}.

Define $T_{\delta} = \inf\left\{T \in \mN: \sqrt{T} + \frac{c(t,\delta)}{C_{\epsilon}(M)} \le T \right\}$. Hence, we have: 
\[
\mE\left[\tau_{\delta}\right] \le T_{\epsilon} + T_{\delta} + \sum_{T=1}^{\infty} 2 K^2 \frac{1}{1-\exp(-C)} T \exp\left(-CT^{-1/8}\right) \le T_{\epsilon} + T_{\delta} + T'
\]
Let $C(\eta) = \inf\{T: T -\sqrt{T} \ge \frac{T}{(1+\eta)}\}$. Then: 
\[
T_{\delta} \le C(\eta) + \inf \left\{ T \in \mN: \frac{T C_{\epsilon}(M)}{(1+\eta)} \ge \beta(t,\delta)\right\}
\]

\medskip
\textbf{Step 6: Obtaining the asymptotic bounds}
\medskip 
Taking limits: 
\[
\lim_{\delta \rightarrow 0} \inf \frac{\mE\left[\tau_{\delta}\right]}{\log\left(\frac{1}{\delta}\right)} \le \alpha \cT(M) \ \forall \ \alpha \ge 1
\]
\end{proof}

\newpage
\section{Reduction to Best-arm Identification}
We briefly discuss how the metric $\pdist{\cdot}{\cdot}$ extends existing notions of gap in best-arm and Pareto-front identification literature. Specifically, we proceed with the three following observations.

\begin{enumerate}[leftmargin=*]
\item Observe that $\pdist{\pftrue}{\pftrue} = 0$. \item Further, iff $\cone$ represents the component-wise ordering as in Pareto-front identification~\citep{auer2016pareto,kone2023adaptive}, then $z^{(\ell)} = 1, \ \forall \ \ell \in [L]$.
\begin{align*}
&\pdist{\armset\setminus \pftrue}{\pftrue})\\
&= \max\left\{ \sup_{k \in \armset \setminus \pftrue} d\left(k,\pftrue \right) , \sup_{k \in \pftrue} d\left(\armset\setminus\pftrue, k \right)\right\} \\
&= \max\Bigg\{ \sup_{k \in \armset\setminus \pftrue} \inf_{k' \in \pftrue} \max\left\{0,\min_{\ell \in [L]} \left(\mutrue{\ell}{k}-\mutrue{\ell}{k'}\right)\right\},\\
&\quad\quad\quad\quad\quad \sup_{k \in \pftrue}\inf_{k' \in \armset\setminus \pftrue} \max\left\{0,\min_{\ell \in [L]} \left(\mutrue{\ell}{k}-\mutrue{\ell}{k'}\right)\right\} \Bigg\}\\
&= \max\Bigg\{ \sup_{k \in \armset\setminus \pftrue} \inf_{k' \in \pftrue} \max\left\{0,m(k',k)\right\}, \\
&\quad\quad\quad\quad\quad \sup_{k \in \pftrue}\inf_{k' \in \armset\setminus \pftrue} \max\left\{0,M(k,k')\right\}\Bigg\} \\
&= \sup_{k \in \pftrue}\inf_{k' \in \armset\setminus \pftrue} \max\left\{0,M(k,k')\right\} \,.
\end{align*}
\item Finally, when there is only a single objective, i.e., $\vert L \vert=1$ and assuming an unique optimal arm (fairly common assumption in BAI literature), we have:
\begin{align*}
\pdist{\armset\setminus \pftrue}{\pftrue}) &= \max\left\{ \sup_{k \in \armset \setminus \pftrue} d\left(k,\pftrue \right) , \sup_{k \in \pftrue} d\left(\armset\setminus\pftrue, k \right)\right\} \\
&= \max\Bigg\{ \sup_{k \in \armset\setminus k^{\ast}} \max\left\{0, \left(\mu_{k}-\mu_{k^{\ast}}\right)\right\}, \\
&\quad\quad\quad\quad\quad \sup_{k' \in \armset\setminus k^{\ast}} \max\left\{0, \left(\mu_{k^{\ast}}-\mu_{k'}\right)\right\} \Bigg\} \\
&= \min_{k' \neq k^{\ast}} \Delta_{k'}\,,
\end{align*}
which is exactly the gap for one-dimensional bandit.
\end{enumerate}
\clearpage\section{Useful Existing Results}
\label{sec:aux}

\begin{theorem}[Berge's Maximum Theorem~\citep{berge1877topological}]
\label{thm:berge}
Let $\cU$ and $\cV$ be topological spaces, $f: \cU \times \cV \rightarrow \mR$ and $C: \cU \rightarrow \cV$ be non-empty compact set for all $u \in \cU$. Then, if $C$ is continuous at $u$, $f^\ast(u) = \max_{v \in C(u)} f(u,v)$ is continuous and $C^\ast(u) = \{v \in C(u) : f^\ast(u) = f(u,v)\}$ is upper-hemicontinuous. 
\end{theorem}

\begin{theorem}[Donsker-Vardhan Variational Formula~\citep{donsker1975asymptotic}]
\label{thm:donsker-vardhan}
For mutual information $\kl{P}{Q}$, we have that: 
\[
\kl{P}{Q} = \sup_{f} \mE_{P} \left[f\right] - \ln \mE_{Q} \left[ \exp(f) \right]
\]
\end{theorem}

\begin{lemma}[Peskun Ordering~\citep{peskun1973optimum}]
\label{lem:peskun-order}
For any two random variables $X,Y$ on $\mR^{K}$ the following are equivalent: 
\begin{enumerate}
\item $X \le_{s} Y$
\item For all $x \in \mR^{K}, \ \mP\left[ X \ge x\right] \le \mP\left[ Y \ge x\right]$
\item For all non-negative functions $f_{1},f_{2},\ldots,f_{K}$, we have that: $ \mathbb{E}[\Pi_{i=1}^{K} f_{i}]  \le \mathbb{E}[\Pi_{i=1}^{K} f_{i}]$
\end{enumerate}
\end{lemma}

\begin{lemma}[\cite{magureanu2014lipschitz}]
\label{lem:kl-exp-bound}
For any $k=1,2,\ldots,K$, let us define $1\le t_k \le t$. Then for all $0 \le C_{k} \le M_{k}$, we have 
\[
\mP\left(\cap_{k \in [K]} \left\{ \hat{M}_{k,t} \le C_{k},\ t_{k} \le N_{k,t}\right\} \right) \le \exp\left( -\sum_{k \in [K]}t_k\kl{C_k}{M_k}\right)\,.
\]
\end{lemma}

\begin{lemma}[\cite{magureanu2014lipschitz}]
\label{lem:multidim-exp-sum}
Let $a > 0$ and $K \ge 2$ and $Z \in \mR^{K}$ such that for all $\xi \in \mR^{K}_{+}$ we have:
\[
\mP\left( Z  \ge  \zeta \right) \le \exp\left(-a\sum_{k \in [K]} \zeta_{k} \right)\,.
\]
Then, for all $\rho \ge \frac{K}{a} \in \mR_{+}$, we have
\[
\mP\left( \sum_{k \in [K]} Z_{k} \ge  \rho\right) \le \left(\frac{ae\rho}{K}\right)^{K} \exp(-a\rho)\,.
\]
\end{lemma}

\begin{lemma}[Single-arm concentration~\cite{kaufmann2021mixture}]
\label{lem:single-arm-conc}
The following is a $\delta$ uniformly valid confidence interval on $z^{\top}\mu$: 
\[
z^{\top}\mu \in \left[z^{\top}\hat{\mu}+\sqrt{2\left(LC^{g}\left(\frac{\ln(\frac{1}{\delta})}{L}\right)\right) +\sum_{k \in [K]}c \ln \left( d + \ln \ln(N_{k,t}) \right)\frac{\sum_{\ell \in L}(z^{\ell})^{2}}{N_{k,t}}}\right]
\]
\end{lemma}



\newpage
\section*{NeurIPS Paper Checklist}

The checklist is designed to encourage best practices for responsible machine learning research, addressing issues of reproducibility, transparency, research ethics, and societal impact. Do not remove the checklist: {\bf The papers not including the checklist will be desk rejected.} The checklist should follow the references and precede the (optional) supplemental material.  The checklist does NOT count towards the page
limit. 

Please read the checklist guidelines carefully for information on how to answer these questions. For each question in the checklist:
\begin{itemize}
    \item You should answer \answerYes{}, \answerNo{}, or \answerNA{}.
    \item \answerNA{} means either that the question is Not Applicable for that particular paper or the relevant information is Not Available.
    \item Please provide a short (1–2 sentence) justification right after your answer (even for NA). 
\end{itemize}

{\bf The checklist answers are an integral part of your paper submission.} They are visible to the reviewers, area chairs, senior area chairs, and ethics reviewers. You will be asked to also include it (after eventual revisions) with the final version of your paper, and its final version will be published with the paper.

The reviewers of your paper will be asked to use the checklist as one of the factors in their evaluation. While "\answerYes{}" is generally preferable to "\answerNo{}", it is perfectly acceptable to answer "\answerNo{}" provided a proper justification is given (e.g., "error bars are not reported because it would be too computationally expensive" or "we were unable to find the license for the dataset we used"). In general, answering "\answerNo{}" or "\answerNA{}" is not grounds for rejection. While the questions are phrased in a binary way, we acknowledge that the true answer is often more nuanced, so please just use your best judgment and write a justification to elaborate. All supporting evidence can appear either in the main paper or the supplemental material, provided in appendix. If you answer \answerYes{} to a question, in the justification please point to the section(s) where related material for the question can be found.

IMPORTANT, please:
\begin{itemize}
    \item {\bf Delete this instruction block, but keep the section heading ``NeurIPS paper checklist"},
    \item  {\bf Keep the checklist subsection headings, questions/answers and guidelines below.}
    \item {\bf Do not modify the questions and only use the provided macros for your answers}.
\end{itemize}


\begin{enumerate}

\item {\bf Claims}
    \item[] Question: Do the main claims made in the abstract and introduction accurately reflect the paper's contributions and scope?
    \item[] Answer: \answerYes{} 
    \item[] Justification: 

\item {\bf Limitations}
    \item[] Question: Does the paper discuss the limitations of the work performed by the authors?
    \item[] Answer: \answerYes{} 
    \item[] Justification: 

\item {\bf Theory Assumptions and Proofs}
    \item[] Question: For each theoretical result, does the paper provide the full set of assumptions and a complete (and correct) proof?
    \item[] Answer: \answerYes{} 
    \item[] Justification: 

    \item {\bf Experimental Result Reproducibility}
    \item[] Question: Does the paper fully disclose all the information needed to reproduce the main experimental results of the paper to the extent that it affects the main claims and/or conclusions of the paper (regardless of whether the code and data are provided or not)?
    \item[] Answer: \answerYes{} 
    \item[] Justification: 

\item {\bf Open access to data and code}
    \item[] Question: Does the paper provide open access to the data and code, with sufficient instructions to faithfully reproduce the main experimental results, as described in supplemental material?
    \item[] Answer: \answerYes{} 
    \item[] Justification: 

\item {\bf Experimental Setting/Details}
    \item[] Question: Does the paper specify all the training and test details (e.g., data splits, hyperparameters, how they were chosen, type of optimizer, etc.) necessary to understand the results?
    \item[] Answer: \answerNA{} 
    \item[] Justification: 

\item {\bf Experiment Statistical Significance}
    \item[] Question: Does the paper report error bars suitably and correctly defined or other appropriate information about the statistical significance of the experiments?
    \item[] Answer: \answerNA{}
    \item[] Justification: 

\item {\bf Experiments Compute Resources}
    \item[] Question: For each experiment, does the paper provide sufficient information on the computer resources (type of compute workers, memory, time of execution) needed to reproduce the experiments?
    \item[] Answer: \answerNA{} 
    \item[] Justification: 
    
\item {\bf Code Of Ethics}
    \item[] Question: Does the research conducted in the paper conform, in every respect, with the NeurIPS Code of Ethics \url{https://neurips.cc/public/EthicsGuidelines}?
    \item[] Answer: \answerYes{} 
    \item[] Justification: 

\item {\bf Broader Impacts}
    \item[] Question: Does the paper discuss both potential positive societal impacts and negative societal impacts of the work performed?
    \item[] Answer: \answerNA{} 
    \item[] Justification: 
    
\item {\bf Safeguards}
    \item[] Question: Does the paper describe safeguards that have been put in place for responsible release of data or models that have a high risk for misuse (e.g., pretrained language models, image generators, or scraped datasets)?
    \item[] Answer: \answerNA{} 
    \item[] Justification:

\item {\bf Licenses for existing assets}
    \item[] Question: Are the creators or original owners of assets (e.g., code, data, models), used in the paper, properly credited and are the license and terms of use explicitly mentioned and properly respected?
    \item[] Answer: \answerNA{} 
    \item[] Justification: 

\item {\bf New Assets}
    \item[] Question: Are new assets introduced in the paper well documented and is the documentation provided alongside the assets?
    \item[] Answer: \answerNA{} 
    \item[] Justification: 

\item {\bf Crowdsourcing and Research with Human Subjects}
    \item[] Question: For crowdsourcing experiments and research with human subjects, does the paper include the full text of instructions given to participants and screenshots, if applicable, as well as details about compensation (if any)? 
    \item[] Answer: \answerNA{} 
    \item[] Justification: 

\item {\bf Institutional Review Board (IRB) Approvals or Equivalent for Research with Human Subjects}
    \item[] Question: Does the paper describe potential risks incurred by study participants, whether such risks were disclosed to the subjects, and whether Institutional Review Board (IRB) approvals (or an equivalent approval/review based on the requirements of your country or institution) were obtained?
    \item[] Answer: \answerNA{} 
    \item[] Justification: 

\end{enumerate}

\end{document}